\documentclass[9pt,a4paper,twocolumn,twoside]{rho-class/rho}
\usepackage[english]{babel}
\usepackage{tabularx}

\title{Evaluating Multimodal LLMs for Inpatient Diagnosis: Real-World Performance, Safety, and Cost Across Ten Frontier Models}

\author[{\textasteriskcentered}, a,b,c]{Bruce A. Bassett}
\author[b]{Amy Rouillard}
\author[c,d]{Sitwala Mundia}
\author[e]{Michael Cameron Gramanie}
\author[c]{Linda Camara}
\author[e]{Ziyaad Dangor}
\author[e]{Shabir A. Madhi}
\author[e]{Kajal Morar}
\author[e]{Marlvin T. Ncube}
\author[$\dagger$,f]{Ismail Kalla}
\author[$\dagger$, g]{Haroon Saloojee}

\affil[a]{School of Computer Science and Applied Mathematics, University of the Witwatersrand, Johannesburg, South Africa}
\affil[b]{Wits MIND Institute, University of the Witwatersrand, Johannesburg, South Africa}
\affil[c]{Grai Labs, Cape Town, South Africa}
\affil[d]{Faculty of Health Sciences, University of the Witwatersrand, Johannesburg, South Africa}
\affil[e]{
South African Medical Research Council Vaccines and Infectious Diseases Analytics Research Unit, Faculty of Health Sciences, University of the Witwatersrand, Johannesburg, South Africa}
\affil[f]{Department of Internal Medicine, Charlotte Maxeke Johannesburg Academic Hospital, South Africa }
\affil[g]{Department of Paediatrics and Child Health, Faculty of Health Sciences, University of the Witwatersrand, Johannesburg, South Africa}

\dates{\today}

\corres{\textsuperscript{\textasteriskcentered}Corresponding author: \href{mailto:author.one@institute.org}{bruce.a.bassett@gmail.com} }
\corres{\textsuperscript{\textdagger} Joint senior authors.}

\begin{abstract}

{\bf Background:} Large language models (LLMs) are increasingly proposed for diagnostic decision support, yet few evaluations use real-world multimodal inpatient data, particularly in low- and middle-income country (LMIC) public hospitals.

\vspace{0.4em}

{\bf Methods:} We conducted VALID, a retrospective diagnostic-accuracy evaluation of 539 multimodal inpatient cases from a tertiary public hospital in South Africa. Multimodal inputs included radiology (CT, MRI, X-ray, ultrasound) and radiology reports supplied as images. Laboratory results, clinical narratives and vital signs were supplied as structured text. Expert panels independently reviewed and adjudicated 300 cases split between a Balanced, random set of 139 cases and a Discordant subset of 161 cases selected for maximal disagreement between ward and ensemble LLM model diagnoses. These 300 form reference ground truth diagnoses, differentials, and clinical reasoning. Ten leading multimodal LLMs generated diagnostic outputs under zero-shot, closed-book conditions. A panel-calibrated LLM Jury of three independent frontier models scored all model outputs and routine ward diagnoses against the expert reference standards across four dimensions: diagnostic accuracy, differential diagnosis quality, clinical reasoning quality, and patient-safety impact, leading to over 10,000 model evaluations. Routine ward diagnoses served as a real-world clinical comparator. A subset of 20 cases where expert panels and models maximally disagreed was reviewed by senior tie-breaker panels. The primary outcomes were weighted composite scores across the dimensions ($S_3$ and $S_4$) and case-by-case win-rate.

\vspace{0.4em}

{\bf Results:} We found that (i)~across all ten LLMs, mean diagnostic performance was tightly clustered, with $<15\%$ relative variation in scores despite a 50x range in cost. GPT-5 Mini's performance was within 5\% of the top model for a mean cost of just \$0.02 per case. (ii)~All LLMs achieved significantly higher average diagnosis and treatment-safety scores than routine ward diagnoses (at 95\% CI). Rates of low patient safety scores for models were comparable to or smaller than those for the routine ward diagnoses. (iii)~GPT-5.1 at high reasoning scored best across all metrics, followed closely by Gemini-2.5 and Gemini-3-preview.  (iv)~The tie-breaker cases evaluated by the most senior panels provide evidence that LLM ensemble diagnoses outperformed the expert panels. (v)~Access to the radiological reports in addition to imaging improved LLM scores on average by $\simeq 6\%$. (vi)~  LLM scores for diagnosis and clinical reasoning are highly correlated (Spearman $\rho = 0.85$).  (vii)~The rates at which LLMs provide a diagnosis varied between $65\%$ (Anthropic) and $100\%$ (Google), driven primarily by model context image limits. (viii)~All results were stable to choice of panel subsets (Balanced vs Discordant) and to all details of the LLM Jury which proved stable and reliable across all tests.

\vspace{0.4em}
{\bf Conclusions:} We evaluated 10 leading multimodal LLMs on a challenging, real-world LMIC public-hospital inpatient dataset using expert adjudication as ground truth. The LLMs demonstrated broadly similar diagnostic performance despite large differences in cost. LLMs outperformed ward diagnoses on average treatment-safety metrics, though tail-risk recommendations remain a concern. For LMIC health systems, considerations such as affordability, input constraints, robustness, and governance may be as important as marginal differences in mean performance.

\vspace{0.4em}

{\bf Clinical Relevance:} Multimodal LLMs may provide useful, highly cost-effective diagnostic decision support in high-volume LMIC public-sector hospitals where LLM cost per case is significantly cheaper than specialists, even in LMICs. 
\end{abstract}


\begin{document}

    \maketitle
    \thispagestyle{firststyle}


\section{Introduction}
\rhostart{D}iagnostic error is a major source of preventable harm worldwide, with amplified effects in health systems facing high patient volumes, limited specialist availability, and constrained access to diagnostic investigations~\cite{Balogh2015ImprovingDiagnosis, who_diagnostic_errors_2016}. These challenges are particularly prominent in public-sector hospitals in low- and middle-income countries (LMICs), where multimodal clinical information, such as laboratory tests or imaging, may be incomplete, fragmented, or inconsistently documented~\cite{who_diagnostic_errors_2016}. Evaluating clinical AI tools in such environments is essential before they can be responsibly deployed.

Large language models (LLMs), now increasingly capable of processing complex multimodal inputs, have been proposed as potential assistants for diagnostic reasoning, summarisation, and identification of safety concerns \cite{Omiye_2024, Saab2024GeminiMedicine}. While LLMs have been shown to outperform medical professionals on some tasks \cite{rutunda2026_frontline_low_resource_ref7,vanveen2024_clinical_summarization,nori2025sequential}, most contemporary evaluations rely on synthetic vignettes, static question-answer pairs, or single-modality inputs that do not reflect the heterogeneity or complexity of real inpatient care \cite{qazi2026_llm_diagnostic_assistance_lmic,Xu2025DiagnosisTriage,Reese2023LimitationsClinicalDiagnosis, artsi2025_real_world_workflows_systematic_review}. There are recent attempts to move towards realistic conditions \cite{brodeur2026prospectiveclinicalfeasibilitystudy, yao2025_realworld_diagnosis_prompting, agweyu2026_kenya_emr_embedded_llm_safety} but few studies directly compare LLMs with routine ward diagnoses -- the actual diagnostic decisions made under real working conditions \cite{artsi2025_real_world_workflows_systematic_review}. Even fewer incorporate expert clinical consensus as a reference standard while simultaneously evaluating a diverse set of multimodal LLMs in a real LMIC context \cite{rutunda2026_frontline_low_resource_ref7}.

This study addresses these gaps by evaluating ten frontier multimodal LLMs (as of October 2025) using adult and paediatric inpatient records from a high-volume South African tertiary public hospital. Each patient record included multimodal artefacts -- clinical narratives, laboratory results, radiology reports and imaging -- reflecting the real diagnostic information available during clinical care. To establish robust reference standards, specialist expert panels independently reviewed and adjudicated 300 records, producing consensus diagnoses and differential diagnoses, and explaining clinical reasoning. Ward diagnoses recorded during patient care were included as a real-world comparator.

To enable consistent scoring at scale, we used a calibrated “LLM Jury”, an idea that has shown benefits in previous studies~\cite{williams2025human,chiang_can_2023, croxford_evaluating_2025, croxford_automating_2025}. Our LLM Jury was composed of three leading models from independent vendors: Anthropic, Google and OpenAI. The LLM Jury evaluated diagnostic accuracy (Dx), quality of differential diagnoses (DDx), clinical reasoning, and patient-safety implications relative to expert reference standards. This approach enabled systematic, reproducible, case-level comparisons between LLMs, expert consensus, and ward diagnoses.

The study aimed to answer three key questions:
\begin{enumerate}
    \item How does the diagnostic and patient-safety performance of multimodal LLMs compare with expert consensus reference standards?
    \item How does LLM performance compare with ward diagnoses documented during real-world clinical care?
    \item Do large differences in computational cost between LLMs correspond to meaningful differences in clinical performance?
\end{enumerate}

By addressing these questions in a real LMIC public-hospital cohort with rich multimodal data, this study provides robust evidence to guide the safe, ethical, and effective deployment of LLMs in resource-constrained clinical environments.

\section{Methods}
\subsection{Study Design and Setting}

We conducted a retrospective, cross-sectional diagnostic-accuracy evaluation called VALID (Validation of LLMs versus Doctors in Diagnosis) using real-world multimodal inpatient data from Chris Hani Baragwanath Academic Hospital (CHBAH), a major tertiary public hospital serving a high-volume urban catchment in Johannesburg, South Africa. 

The VALID project is structured not as a single benchmark, but as a linked family of complementary evaluation datasets designed to compare diagnostic performance across tiered expert panels, ward diagnoses, and multiple AI models under varying information and adjudication conditions. This design allows the study to address several distinct but connected questions: how models perform relative to expert consensus, how ward diagnoses compare with panel adjudication, how performance changes when information is degraded and how robust model scoring is across evaluators.

At the core of VALID are two principal case pools. The first is the expert panel-reviewed cohort (300 records), for which expert panel diagnoses provide a high-quality reference standard. Around this reference, the study defines a set of targeted subsets that enable secondary analyses such as re-scoring, tie-break adjudication, and reproducibility testing. This creates a design in which the same cases are repeatedly used in different configurations, allowing each dataset to answer a specific methodological question while also reinforcing the others.

The VALID framework varies multiple evaluation axes in a controlled, combinatorial manner. These axes include:
\begin{enumerate}
    \item The case subset being analysed.
    \item The source of ground truth, which may be expert panel diagnosis, ward diagnosis, or a senior tie-break panel diagnosis.
    \item The diagnosis being scored, which may be expert panel diagnosis, ward diagnosis, or AI model diagnosis.
    \item The scoring authority, either a human panel or the LLM Jury.
\end{enumerate}

\subsection{Data} 
\subsubsection{Case sources, time window, and wards}
Data collection commenced in April 2025, using retrospectively sourced records from inpatients discharged March–October 2025. Approximately 80\% of cases were drawn from Internal Medicine and 20\% from Paediatrics. A total of 539 patient records were abstracted. Age and sex distributions were broad and reflected routine inpatient demographics at the study hospital.

Central nervous system and respiratory presentations comprised the largest categories in the full dataset, followed by cardiovascular conditions, Table~\eqref{tab:case_distribution}. This distribution mirrors routine ward presentation patterns and the study’s multimodality inclusion requirement. 
Model outputs and expert adjudications were based on the same retrospectively assembled admission record, with no additional information introduced between index and reference. All model diagnoses and LLM Jury evaluations were made in October 2025 for consistency. 

Ward diagnoses were extracted from routine clinical documentation and used as a real-world human comparison baseline. Ward notes commonly lacked structured differential diagnoses and explicit reasoning. Ward outputs were evaluated with the same instrument wherever comparable with re-weighted composites to account for missing reasoning, see Section~\eqref{sec:outcomes-and-scoring}. 

\begin{table}[htbp]
\centering
\renewcommand{\arraystretch}{1.2}
\begin{tabular}{lccc}
\hline
\textbf{System} & \textbf{Total n (\%)} & \textbf{Adult n (\%)} & \textbf{Paediatric n (\%)} \\
\hline
Central Nervous System & 178 (33) & 157 (37) & 21 (18) \\
Respiratory & 129 (24) & 89 (21) & 40 (35) \\
Cardiovascular & 73 (14) & 67 (16) & 6 (5) \\
Gastrointestinal & 44 (8) & 32 (8) & 12 (11) \\
Renal & 36 (7) & 27 (6) & 9 (8) \\
Oncology & 28 (5) & 14 (3) & 14 (12) \\
Other & 17 (3) & 11 (3) & 6 (5) \\
Rheumatology & 11 (2) & 9 (2) & 2 (2) \\
Haematology & 9 (2) & 8 (2) & 1 (1) \\
Endocrine & 9 (2) & 8 (2) & 1 (1) \\
Psychiatric & 2 (0) & 2 (0) & 0 (0) \\
Dermatology & 2 (0) & 0 (0) & 2 (2) \\
Musculoskeletal & 1 (0) & 1 (0) & 0 (0) \\
\hline
\end{tabular}
\vspace{1em}
\caption{Distribution of the full abstracted cohort (n=539) by clinical system, with corresponding adult and paediatric case counts and percentages.}
\label{tab:case_distribution}
\end{table}

\begin{figure}
    \centering
    \includegraphics[width=\linewidth]{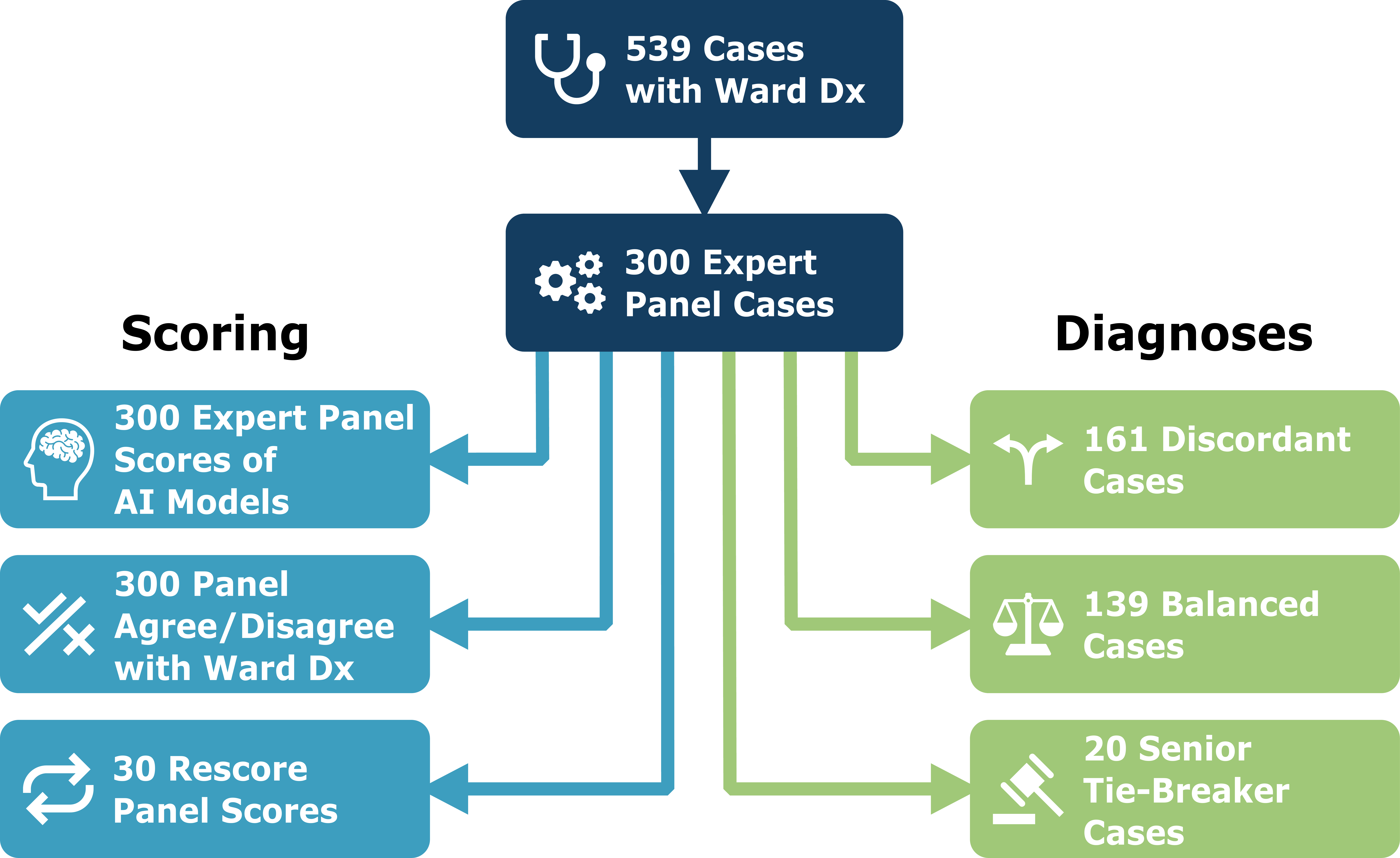}
    \caption{Schematic illustration of the outputs from the expert panels that adjudicated 300 of the full 539 cases.  Panels provided diagnoses and clinical reasoning for the Balanced, Discordant, and tie-breaker subsets (right). For each case, the panels also gave a binary agree/disagree decision with the ward's Dx and scored a randomly-selected LLM Dx for LLM Jury calibration purposes. Panels also re-scored 30 cases to evaluate inter-rater reliability (left). The precise panel workflow is shown in Figure~\eqref{fig:eval-workflow}.}
    \label{fig:panel-outputs}
\end{figure}

\subsubsection{Eligibility criteria and data requirements}

\hspace*{1em}\textbf{Inclusion.} Patient records required laboratory results and at least one additional clinically relevant diagnostic modality obtained during the index admission (i.e. ECG, chest radiography, CT, MRI, or ultrasound).

\textbf{Exclusion.} We excluded cases with insufficient diagnostic complexity, missing core clinical information (e.g., missing presenting complaint/history/examination), or where essential investigations for the suspected condition were unavailable (e.g., suspected pneumonia without a chest radiograph; suspected stroke without brain imaging). Relevant prior-admission investigations were excluded unless clearly tied to the index presentation. Post-discharge outpatient imaging was excluded. Missing medication/social histories were not exclusionary if the principal narrative and findings were present.

\begin{figure}
    \centering
    \includegraphics[width=\linewidth]{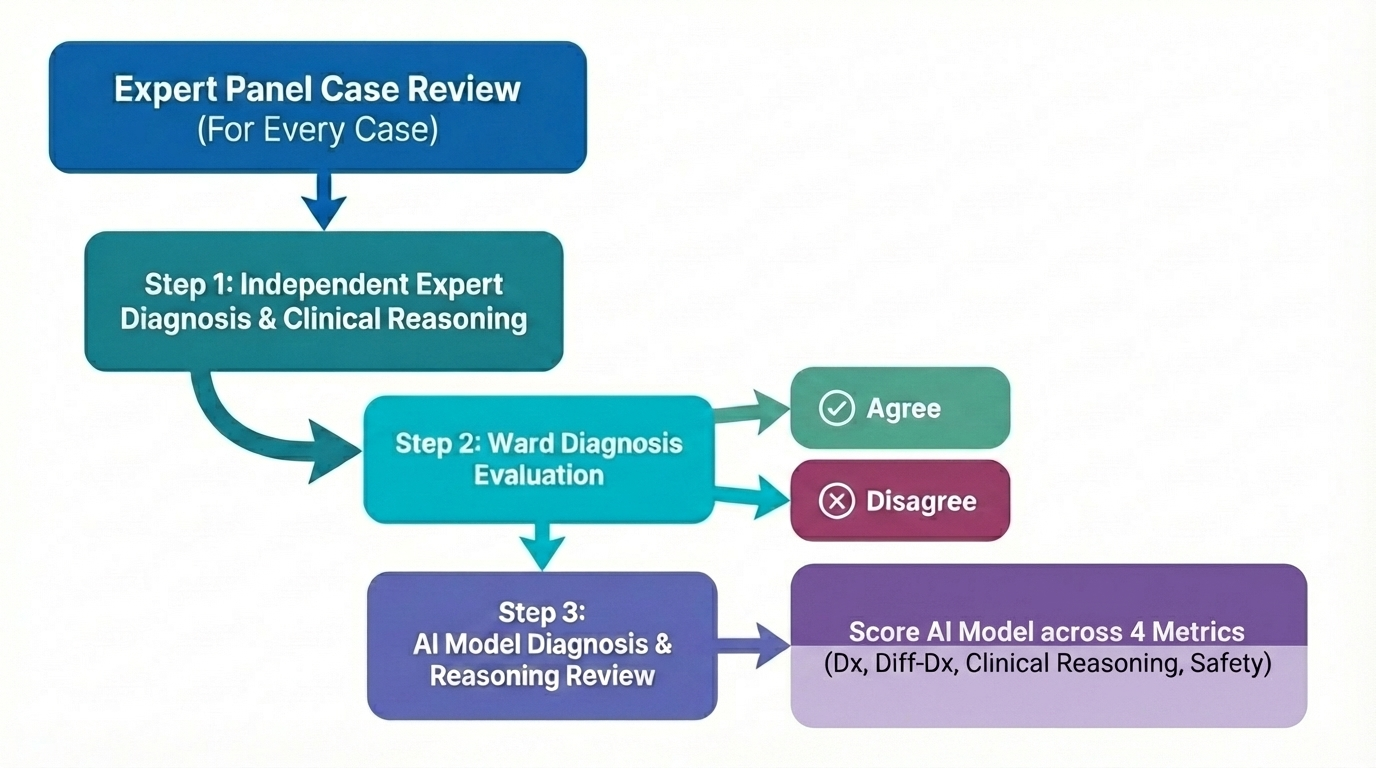}
    \caption{The expert panel evaluation workflow followed in the adjudication of the 300 combined Balanced and Discordant cases.}
    \label{fig:eval-workflow}
\end{figure}

\subsubsection{Data Capture and Quality Assurance}
Data were abstracted from: (1) physical inpatient records, (2) digital laboratory systems, and (3) radiology reports and imaging archives. Clinical content prioritised registrar/consultant notes and was transcribed without interpretive “clean-up” to preserve fidelity. Study physician-led quality assurance (QA) included early review of new abstractors’ initial cases and physician proofreading of all cases, with targeted verification against source records when concerns were identified. 

\subsubsection{Multimodal Inputs and Pre-processing}

Each case included clinical narratives, structured vital signs, laboratory results, radiology reports, and imaging. PDF reports were converted to images for ingestion. If DICOM files had more than 90 frames, they were uniformly downsampled to 90-frame videos with default windowing choices. This number traded spatial slice resolution (not more than 10mm gaps between images) against maximising the number of returned diagnoses from the Claude models (which had a 100-image maximum limit). Summary statistics of the distribution of the number of images per patient record are shown in Table~\eqref{fig:summary_stats}. The largest case had over 1200 associated images, mainly derived from CT and/or MRI scans.

\subsubsection{Case selection for expert panel review}

Of the 539 available cases, 300 cases were selected for expert panel adjudication. This number was determined pragmatically based on time and cost constraints. A sample of 300 cases yields narrow bootstrap confidence intervals for group‑level comparison of LLM performance, approximately 3\% uncertainty (95\% CI) on mean model scores. 

The 300 cases chosen for expert panel adjudication were formed from two subsets:
\begin{itemize}
    \item \textbf{Balanced Subset:} 139 of the 300 cases were selected by random sampling from the available pool of abstracted cases at the time, with pragmatic balancing across diagnosis categories for diversity. These cases were evaluated by the panels first. 
    \item \textbf{Discordant Subset:} 161 of the 300 cases were selected based on maximal diagnostic disagreement between ward diagnoses and the LLM outputs. Disagreement was estimated by using the average patient safety score using the ward diagnosis as ground truth. \footnote{This approach had the advantage of not requiring models to have the same diagnosis as other models in order to flag disagreement with the ward. As long as most models disagreed with the ward diagnosis, the average safety score would be low.} The goal was to improve the health of the full dataset and to create a challenge dataset where either the ward or the LLMs had likely made a mistake, identified by substantial disagreement between the LLM and the ward diagnoses. 
\end{itemize}

\subsection{Expert Panels}
\subsubsection{Composition and Workflow}

Expert panels were composed of two senior internal medicine or two paediatric specialists. Consensus adjudication by two specialists was selected because no single gold standard applies across the heterogeneous inpatient conditions evaluated. Expert consensus reflects real‑world diagnostic reasoning and supports consistent case‑level reference outputs. Panels had access to both the radiology reports and videos created from the DICOM files, reflecting the image data used for LLM diagnosis. 

Cases were first reviewed independently by panellists who then met to discuss the cases. The panel workflow \footnote{Panels used a purpose-build evaluation platform that gave them access to all patient records and captured their diagnoses and evaluations.}, see Figures~\eqref{fig:panel-outputs} and \eqref{fig:eval-workflow}, consisted of (i) coming to a consensus on a combined primary and secondary diagnosis, differential diagnosis at admission and clinical reasoning for the combined diagnosis. (ii) The panel then reviewed the routine ward diagnosis and differential, either agreeing or disagreeing, and (iii) finally scored a single randomly selected LLM diagnosis and clinical reasoning for that case across all dimensions on a 1-5 Likert scale, scoring detailed in Section~\eqref{sec:outcomes-and-scoring}.

This provided a set of 330 expert panel evaluations, $300$ of 23 different LLMs and $30$ of the ward, which was used for calibrating the LLM Jury and testing inter-expert and panel-LLM-Jury agreement, see Section~\ref{sec:llm_jury}.

\subsubsection{Senior Tie-Breaker Panels}
\label{sec:tie-breaker}
The two most senior expert panels (the “tie-breaker panels”) reevaluated 20 cases already seen by expert panels: 10 from internal medicine and 10 from paediatrics. The selected cases had maximal diagnostic disagreement between the expert panel diagnoses and the consensus LLM diagnoses (analogous to how the \textbf{\textit{Discordant}} dataset was selected above). Discordance between the LLMs and the expert panels could be explained by two possibilities: (1) the majority of the LLMs had incorrect diagnoses, and potentially an important communal blind spot or LMIC bias, or (2) the expert panel diagnoses were incorrect. These senior tie-breaker panels had access to both the LLM diagnoses and expert panel diagnoses, and their role was to break the deadlock between humans and AI. 

\subsubsection{Re-score Panels}

Finally, to estimate inter-rater reliability between human panels for comparison with the LLM Jury, 30 of the 300 AI model diagnoses were randomly selected to be re-scored by a separate expert human panel. This panel most closely mimicked the LLM Jury conditions. Namely, they received the ground-truth expert panel diagnoses and clinical reasoning, and the corresponding outputs from the selected LLM, which they scored on the same 1-5 Likert scale as all other evaluators. The re-score panels did not receive any case information. 

\subsection{Multimodal Model Choice} 

Choice of the final ten multimodal LLMs to evaluate was delayed until October 2025. At that time, models were chosen spanning a wide cost range from four leading commercial LLM providers: Anthropic, Google, OpenAI and xAI. 

Table~\eqref{tab:model_card} provides specifics of models chosen, versions, costs and model limitations. All models were accessed via commercial APIs under closed-book conditions (no tools or retrieval), and one output per case was generated. If an API failed to return a response, we allowed up to five retries if the failure was due to high demand. All models received the same multimodal artefacts. 

\textbf{No Model Training or Fine‑Tuning.} All LLMs were evaluated in their commercial, pre‑trained form. No model was trained, fine‑tuned, or adapted using study data. All outputs represent zero-shot closed‑book inference under standard API conditions. All parameter choices were set to default values determined by vendors (e.g. temperature). The level of reasoning effort was not specified in any model other than GPT-5.1 which was set to "High". In the case of Gemini 3 Pro (Preview) this unspecified default corresponded to "dynamic" reasoning. 

\textbf{Prompt Optimisation.} With the goal of optimising results, we systematically explored a total of six versions of the diagnostic prompts over the course of data collection. All models were given the same final prompt, i.e. no per-model prompt optimisation was undertaken.

\subsection{Cost per case vs Cost per Million Tokens}

Most models processed tokens similarly. The exception was Gemini 3 Pro (Preview), which used many more image tokens than other models to process cases. The cost-per-million-tokens varied from \$0.2 (\$0.5) to \$15 (\$75) for input (output) tokens (a 75-150x variation) while the cost-per-case, which combines input and output costs, only varied by a factor of 50x.

\subsection{Reproducibility} 

Because the index tests were commercial models accessed via API, full model weights and training details were unavailable. Reproducibility is supported by (1) identical multimodal inputs provided to all models, (2) use of standard, fixed prompts for both diagnostic requests and LLM Jury requests, (3) closed‑book conditions, (4) one‑pass API calls with default model parameters, (5) full documentation of input preprocessing workflows and (6) use of specific, fixed models versions with all API calls performed within four weeks in October 2025. In more detail:
\begin{itemize}
    \item \textbf{Closed-book evaluation:} no tool use or external knowledge sources were allowed for LLM diagnoses. 
    \item \textbf{One-pass generation:} one output per model–case pair was generated using a standardised prompt across models. 
    \item \textbf{Ward-baseline comparability:} to avoid imputing missing reasoning, a composite re-weighted $S_3$ score across the common metrics, introduced in Section~\eqref{sec:outcomes-and-scoring}, is used for comparison.  
    \item \textbf{Calibration disclosure:} isotonic regression was used to calibrate LLM Jury scores to expert panels; both raw and calibrated results are reported. The associated paper, \cite{Rouillard2026LLMJury}, gives detailed evaluations of the LLM Jury and calibration. 
\end{itemize}

\subsection{Handling of Missing-ness and Non-responses}

If an API call timed out or did not return a response for any reason other than reaching image context limits, the model was allowed up to 5 attempts, after which the diagnosis attempt was deemed failed. Model non-responses due to failure or vendor limits were treated as missing. For example, Claude models’ 100-image cap reduced response rates to 65-66\%, while Gemini models achieved  99-100\% response rates. Analyses were performed on available cases per model, with case counts reported in \cite{Rouillard2026LLMJury}. No imputation was performed for missing outputs.

\begin{table}[htbp]
\centering
\renewcommand{\arraystretch}{1.2}
\begin{tabular}{p{0.2\linewidth} p{0.3\linewidth} p{0.2\linewidth}}
\hline
\textbf{Statistic} & \textbf{Expert Panel Subset} & \textbf{Full Sample} \\
\hline

Min & 1 & 1 \\
Max & 1272 & 1272 \\
Mean & 103 & 121 \\
Median & 91 & 92 \\

\hline
\vspace{1em}
\end{tabular}
\caption{Summary statistics of the number of images per case for the full sample (n = 539) and the Expert-Panel sample (n = 300). Although the medians are very similar, seven cases with more than 600 images are present in the full sample but not in the expert-panel subset, raising the mean from 103 to 121.}
\label{fig:summary_stats}
\end{table}

\begin{table*}[htbp]
\renewcommand{\arraystretch}{1.2}
\centering

\begin{tabular}{p{0.07\linewidth} p{0.16\linewidth} p{0.08\linewidth} p{0.09\linewidth} p{0.09\linewidth} p{0.07\linewidth} p{0.07\linewidth} p{0.07\linewidth} p{0.08\linewidth}}
\hline
\textbf{Vendor} & \textbf{Model} & \textbf{Variant} & \textbf{Reasoning Level} & \textbf{Max Context Length} & \textbf{Size Limit} & \textbf{Image Limit} & \textbf{Input Cost (USD/MT)} & 
\textbf{Output Cost (USD/MT)} \\ 

\noalign{\vskip 0.4em}
\hline
\noalign{\vskip 0.4em}
Anthropic & Claude 4.1 Opus  & 20250805 & D & 200{,}000 & 32\,MB & 100 & 15 & 75 \\

Anthropic & Claude 4.5 Sonnet & 20250929 & D & 1{,}000{,}000 & 32\,MB & 100 & 3 & 15 \\

Google & Gemini 3 Pro (Preview) & -- & D {\bf (Dynamic)} & 1{,}000{,}000 & 7\,MB/image & 3000 & 2 & 12 \\

Google & Gemini 2.5 Pro & -- & D & 1{,}048{,}576 & 7\,MB/image & 3000 & 1.25 & 10 \\

Google & Gemini 2.5 Flash & -- & D & 1{,}000{,}000 & 7\,MB/image & 3000  & 0.3 & 2.5 \\

OpenAI & GPT-5.1 & 2025-11-13 & {\bf High} & 400{,}000 & 50\,MB & 500  & 1.25 & 10 \\

OpenAI & o3 & 2025-04-16 & D & 200{,}000 & 50\,MB & 500  & 2 & 8 \\

OpenAI & o4-Mini & 2025-04-16 & D & 200{,}000 & 50\,MB & 500  & 1.1 & 4.4 \\

OpenAI & GPT-5 Mini & 2025-08-07 & D & 400{,}000 & 50\,MB & 500  & 0.25 & 2 \\

xAI & Grok 4.1 Fast Reasoning & -- & D & 2{,}000{,}000 & 10\,MB & No limit & 0.2 & 0.5 \\
\noalign{\vskip 0.4em}
\hline
\end{tabular}
\vspace{1em}
\caption{Summary of the ten multimodal LLMs evaluated in this study, ordered by vendor. All LLMs were accessed in October 2025, and costs and limits are as of that date. All models were accessed on {\bf default } ({\bf "D"}, i.e. not specified) parameter values and reasoning effort, apart from GPT-5.1 which was accessed on the highest reasoning effort.  
We show input and output costs (in US\$'s per million tokens) and model constraints (maximum context length in tokens, maximum number of images accepted and total submission size) as of October 2025. Note that the Google models had by far the most permissive constraints in terms of number of images and total memory allowance, which helped ensure nearly 100\% response rate. However, Gemini 3 Pro (Preview) model used significantly more tokens per image than other models. 
}
\label{tab:model_card}
\end{table*}

\subsection{Outcomes and Scoring}\label{sec:outcomes-and-scoring}

All index/comparator outputs were evaluated relative to the expert reference standard across four 1–5 scales: (1) Diagnostic accuracy (combined primary and secondary; Dx). (2) Quality of differential diagnosis at admission (DDx). (3) Clinical reasoning quality, and (4) Patient safety impact, the risk of harm if care followed the index test vs the reference diagnoses, where 1 is very high risk and 5 is minimal expected harm. 

For the diagnosis, we combined the primary and secondary inpatient diagnoses. This was chosen since it was sometimes noted that model primary and secondary diagnoses were flipped relative to the panel, representing a non-essential difference. This reference standard reflects the consensus of two domain‑specialist physicians reviewing all multimodal clinical data from the index admission.

For our primary outcomes, we computed two pre-specified weighted composite scores. The first, denoted $S_4$, combined the four metrics with the following weights:  diagnoses (Dx): 0.30, differential diagnoses (DDx): 0.10, clinical reasoning: 0.30, patient safety: 0.30. Since the captured routine ward diagnoses lacked any diagnostic reasoning we constructed a second score, denoted $S_3$ (reflecting only having three subscores), that excluded reasoning and used weights of (0.40, 0.20, 0.40) for the diagnoses, differential and patient safety scores. A description of each score and the summaries of the weights are shown in Table~\eqref{tab:weights} in the Supplementary. The $S_3$ score allows direct comparison of model and ward scores. Our secondary outcomes include dimension-specific means, safety-tail frequency (proportion with safety $\leq 2$) and case-by-case win-rates.  

While no formal explainability algorithms were applied, the study evaluated the clinical reasoning quality of each model output as a proxy for interpretability. Reasoning quality reflects the coherence, internal justification, and linkage of proposed diagnoses to the multimodal clinical data and offers a structured, clinically meaningful assessment of LLM interpretability.

\subsection{LLM Jury: Composition and Rationale}

Since we had 300 patient records, 10 models to test for each of these and four scores for each model (Dx, DDx, clinical reasoning, patient safety), we needed over 10,000 diagnostic evaluations. It was infeasible to obtain expert panel scores for all our model predictions. Instead, we chose to use an LLM Jury, a method gaining in popularity and credence \cite{Zheng2023LLMJudge, croxford_evaluating_2025}.

The LLM Jury consisted of the following models: Gemini 2.5 Pro, OpenAI o3-2025-04-16, and Anthropic Claude-Opus-4.1. These were chosen from different AI model companies for diversity and for consistency with earlier LLM Jury evaluations (in particular, GPT-5.1 and Gemini 3 only became available at the end of the study).  

The three LLM Jury models independently scored all outputs and operated in a fully blinded manner. They received only the index test output and the reference standard for scoring purposes. Jury models had no access to clinical narratives or expert panel reasoning, and they did not receive any information about model identity, case data or case selection criteria. In addition to the scoring, the LLM Jury was also prompted to provide reasoning for the given scores.

We took the mean of these as the final LLM Jury score for each score and each case. The median offered no clear advantage because of the high agreement between the LLM Jury models. The LLM Jury was compared with, and later calibrated on, 300 expert panel scores of models. Agreement with human panel scoring was generally good, with the LLM Jury consistently stricter than panels. The performance of the LLM Jury is summarised in Section~\eqref{sec:llm_jury} and covered in depth in the associated paper \cite{Rouillard2026LLMJury}.

\subsection{Additional Methodology}
\subsubsection{Statistical Analysis}

We report mean scores with bootstrap confidence intervals (computed from 1,000 samples) and emphasise uncertainty/overlap rather than “winner-takes-all” inference. Models are displayed in ranked order by mean $S_3$ score, interpreted cautiously given tight clustering. We describe the relationship between performance and cost per million tokens using descriptive correlation and visual summaries. No formal hypothesis tests were pre-specified. 

\subsubsection{Ethics and Data Governance}

The study received approval from the University of the Witwatersrand Human Research Ethics Committee, with hospital-level permissions secured prior to data collection. Identifiers were excluded from the analytic dataset. Access was role-restricted, and data were stored on secure devices and South Africa–hosted infrastructure in alignment with POPIA. 

\textbf{Adverse events:} none were associated with the index tests (LLM inference) or with the reference‑standard procedures (expert review of existing records).

\textbf{Deviations from Protocol: }
Not applicable/none reported.

\textbf{Registration:} the study was registered with the South African National Health Research Database (\texttt{GP\_202501\_023}).

\textbf{Protocol availability:} the final protocol is available from the corresponding author upon reasonable request.

\section{Analysis of the LLM Jury Scores}
\label{sec:llm_jury}
In this section, we discuss  our evaluation, testing and calibration of the LLM Jury, summarising key results which are explored in significantly more detail in the accompanying paper \cite{Rouillard2026LLMJury}.

\subsection{Inter-rater reliability}

Overall, the LLM Jury performs well and agreement with expert panels on the overlapping cases-diagnosis pairs is moderate to good but the LLM Jury has better inter-rater ordinal agreement and exhibits better concordance with the primary expert panels than the human re-score panel does. As an example, if one considers the standard Spearman $\rho$ and weighted Cohen $\kappa$ metrics of inter-rater reliability, we find that the LLM Jury aligns better than the re-score panel across all dimensions, as shown in Table~\eqref{tab:rho_kappa1}. Further, the LLM Jury shows excellent agreement with expert panels' rankings and the probability of severe errors is significantly lower in LLM Jury models than in the re-score panel.

\begin{table}[htbp]

\centering
\begin{tabular}{
p{0.07\linewidth}
p{0.22\linewidth} 
    >{\raggedleft\arraybackslash}p{0.1\linewidth}
    >{\raggedleft\arraybackslash}p{0.1\linewidth}
    >{\raggedleft\arraybackslash}p{0.1\linewidth}
    >{\raggedleft\arraybackslash}p{0.1\linewidth}}
\toprule
\textbf{Metric}
&\textbf{Model} 
& \textbf{Dx}
& \textbf{DDx}
& \textbf{Reasoning}
& \textbf{Safety} \\
\midrule

$\rho$ 
\\
& LLM Jury
& {\bf 0.68}
& {\bf 0.52}
& {\bf 0.66}
& {\bf 0.52}
\\
\noalign{\vskip 2pt}
& Re-score Panel
& 0.40
& 0.11
& 0.31
& 0.29
\\

\midrule

$\kappa$ 
\\
& LLM Jury
& {\bf 0.44}
& {\bf 0.22}
& {\bf 0.40}
& {\bf 0.33}
\\

\noalign{\vskip 2pt}

& Re-score Panel
& 0.41
& 0.15
& 0.31
& 0.32
\\

\bottomrule
\end{tabular}
\vspace{1em}
\caption{\textbf{Inter-rater reliability of humans vs LLM Jury:}  Spearman's $\rho$ and Cohen's $\kappa$ for the combined LLM Jury, relative to the primary expert panel scores using the $300$ case diagnoses scored by both the expert panels and LLM Jury and 30 cases scored by the re-score panels. Higher values indicate better agreement, and bold indicates the best-performing evaluators. Across all scores, the LLM Jury aligns more closely with the expert panels than do the (human) re-score panels.}
\label{tab:rho_kappa1}

\end{table}

\subsection{Calibration of LLM Jury scores}

Because the mean LLM Jury scores were about 1 point (on the 1-5 Likert scale) stricter on average than the expert panels (see \cite{Rouillard2026LLMJury}), a natural question is whether results derived from the LLM Jury would be the same if we had been able to use human expert panels for all 10k+ evaluations. To try to address this, we used the 300 expert-panel scores of LLM diagnoses and 30 expert panel evaluations of ward diagnoses to train isotonic regression models.

For each of the three LLM Jury members and for each score (Dx, DDx, clinical reasoning, patient safety), we trained a separate isotonic regression model. These models were tested using 5-fold cross-validation but for the final predictions we used isotonic regressions trained on the full sample of 330 human scores (see \cite{Rouillard2026LLMJury} for details). Isotonic regression is well-suited for calibration in this setting because it makes minimal assumptions about the shape of the mapping between LLm Jury scores and the expert panels' judgments, while enforcing a monotonic relationship that preserves the ordinal structure of the Likert scale.

After calibration, LLM Jury scores had zero mean offset (by construction) relative to the expert panel scores. The mean LLM Jury scores for each model before and after calibration are shown in the supplementary material, Figure~\eqref{fig:Universal-Weighted-Scores}. In tables and figures, we refer to the original (stricter)  LLM Jury scores as {\bf RAW} scores, and the LLM Jury scores calibrated onto the expert panel scores as {\bf ISO} (isotonic). See e.g. Table~\eqref{tab:all_scores}.  

One important note about the isotonic regression is that, because it raises all raw LLM Jury scores, but 5 is still the highest possible score, it compresses the range of scores, as can be seen by comparing values on the left and right sides of Table~\eqref{tab:all_scores}. Whereas there is a 29\% variation from best to worst in the RAW scores, that is reduced to just 13\% after calibration. Figures~\eqref{fig:scores-raw}, \eqref{fig:score-cal}) show the before and after calibration rankings and scores of the LLMs broken down by LLM Jury model. Post isotonic regression, it is clear how the scores have been increased and absolute agreement between LLM Jury models has significantly improved.

\subsection{Consistency of LLM Jury results}

Our main findings are robust to all details of the LLM Jury and calibration. For example, all jury models separately assessed GPT-5.1 as the top-scoring model and all placed the ward diagnoses as weakest, both overall and for the Balanced and Discordant subsets separately. Individual LLM Jury rankings of models between the three jury models were highly concordant (Spearman $\rho$: median 0.92, minimum 0.89). No LLM Jury model showed any systematic preference for diagnoses returned by their model (i.e. model X did not score X's diagnoses preferentially).

When rerunning the LLM Jury models 30 times repeatedly each on 3 case diagnoses, the diagnosis score had a mean coefficient of variation of 0.104, indicative of highly robust and repeatable LLM Jury scores. Variation was not zero since default model temperature values were used and because we demanded an integer score which sometimes caused score ambiguity (e.g. in some cases repeated runs were split equally between scores of 3 and 4). Our analyses, presented extensively in the accompanying paper \cite{Rouillard2026LLMJury}, provide strong evidence that our conclusions are stable to the use of an LLM Jury.

\begin{figure*}
    \centering
    \includegraphics[width=0.8\linewidth]{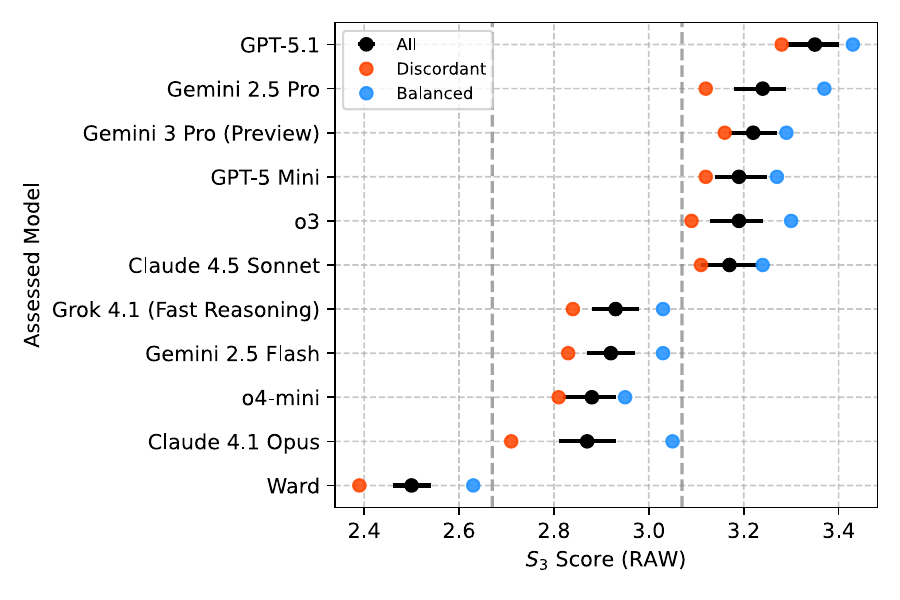}
    \caption{Raw $S_3$ scores per model. $S_3$ combines diagnosis (Dx), clinical reasoning and safety subscores which allows comparison of Ward and LLMs (see Section~\eqref{sec:outcomes-and-scoring}). 68\% confidence intervals are shown on the full dataset (black points/bars). In addition, we show the mean $S_3$ scores on the Balanced (blue) and Discordant samples (red). We see: (1) three statistically significant groups of models separated by the vertical dashed lines. (2) All models are significantly better (at the 95\% confidence level) than the ward scores;  (3) both the ward and all LLMs show significantly reduced performance on the Discordant sample (the biggest drop being for Claude 4.1 Opus). }
    \label{fig:Raw-scores}
\end{figure*}

\section{Results}

\subsection{API responses and missing diagnoses}

While all models received identical multimodal artefacts for each patient record, the response rate of the different models varied dramatically (see the final column in Table~\eqref{tab:all_scores}). For example,  Anthropic's Claude models only gave diagnoses on approximately $65\%$ of cases, primarily due to their 100-image limit per API call, which strongly affected cases with MRIs and CT scans. By contrast, xAI and OpenAI models had approximately $96\%$ response rate, with non-response was either due to exceeding the 10MB case limit (Grok) or the 500 image limit (OpenAI). Google's model response rate exceeded $99\%$. 

All scores are computed using all the available diagnoses for each model, with case counts reported in the final column in Table~\eqref{tab:all_scores}, i.e. we report scores conditioned a diagnosis being returned.  

No imputation was performed for missing model outputs. This means that scores and ranks do {\em not} penalise models for the cases they did not report a diagnosis for, which, for example, would have automatically reduced the Claude scores by approximately one third,  Grok and the OpenAI models by approximately 4\%, and the Google models by less than 1\%. We chose not to penalise models because (a) they are driven by image limits determined in October 2025 by the model providers. Given historical trends, these will likely be significantly relaxed in the next 24 months and (b) the proportion of diagnoses returned for each model is highly sensitive to the amount of MRI and CT imaging, which varies by the sample of patient records chosen.

\subsection{Average Model Scores}

Figure~\eqref{fig:Raw-scores} shows one of the main results: the 10 LLMs ranked by mean RAW $S_3$ Scores (black) with 68\% bootstrap confidence intervals on the means. GPT-5.1 scores best, with the Ward scoring worst. Of the LLMs, Claude 4.1 Opus scores the worst, despite being one of the most expensive. The models are split into three clusters - indicated by the vertical dotted lines - which are significantly separated at 95\% confidence. The Ward diagnoses, scored against the expert panel ground truth form the lowest cluster on their own. The mid-tier cluster is formed from the cheaper models (other than Claude 4.1 Opus), while the top group includes the largest models, other than GPT-5 Mini, which is the only very cheap model in the top group. 

Also shown are the mean scores on the {\bf Balanced} and {\bf Discordant} expert panel subsets (blue and red). Across the board, both the ward and all models did significantly worse on the Discordant subset than the Balanced set, but the drop for the ward was larger than for any of the LLMs, other than for Claude Opus-4.1. This is significant: the Discordant case subset was chosen because the ensemble of LLMs disagreed with the ward diagnosis. The fact that the ward scores dropped more on this subset than did the models, when scored against the expert panel diagnoses, implies that on average the expert panels sided more often with the LLMs than the ward. 

In the Supplementary Tables~\eqref{tab:sub1} and \eqref{tab:sub2}, we break down the individual scores (Dx, DDx, reasoning and safety) for the Balanced and Discordant subsets for both the RAW and ISO calibrated scores. Across both panel-defined subsets, GPT-5.1 is the strongest overall performer, leading the Balanced subset on diagnostic accuracy, differential diagnosis, clinical reasoning, patient safety, and both ISO versions of $S_3$ and $S_4$. Gemini 2.5 Pro is a close second in the Balanced subset, whereas Gemini 3 Pro (Preview) rises to second place in the Discordant subset, suggesting greater robustness when panel disagreement identifies more difficult or ambiguous cases. A consistent pattern is score attenuation in the Discordant subset for all models, with declines of roughly 0.1–0.2 points across most domains, indicating that disagreement-defined cases are systematically harder for both human clinicians and models. Even so, leading LLMs remain clearly above ward performance.

\subsection{Subscore-Specific Performance}

Averages for diagnostic accuracy (Dx), differential diagnosis quality (DDx), clinical reasoning quality, and patient-safety impact are shown in Table~\eqref{tab:all_scores}. Note that since no clinical reasoning was captured for the routine ward diagnoses, there is no clinical reasoning score for the ward.  

GPT-5.1 scored best across all four metrics and in both the raw (RAW) and isotonic calibrated scores (ISO), but the victory margin was small in each case: 0.08, 0.12, 0.03 and 0.07 for RAW Dx, DDx, reasoning and safety scores, respectively. 

\subsection{Performance - Cost Tradeoff}

\begin{figure*}
    \centering
    \includegraphics[width=0.85\linewidth]{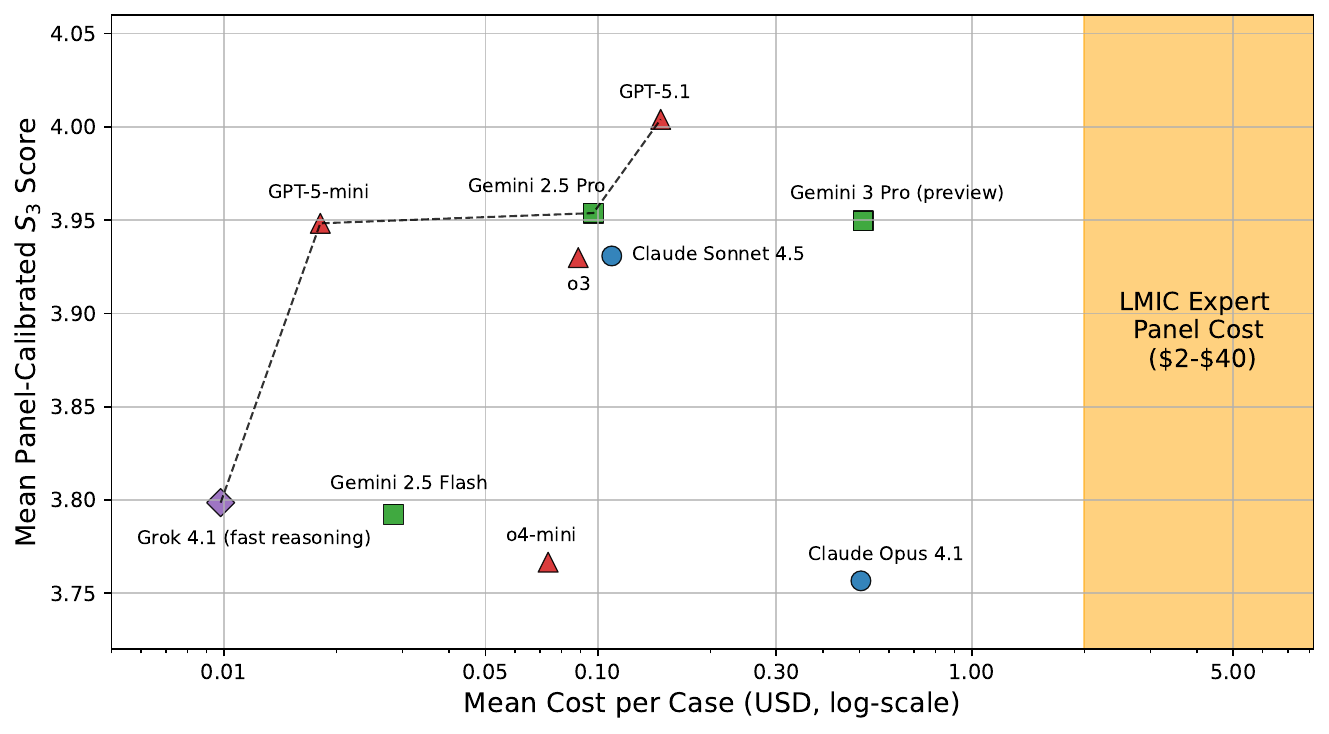}
    \caption{Post-calibration Performance vs mean USD cost per case. The Pareto frontier (dashed line) shows optimal performance as a function of mean model cost per case. Performance between models varied by less than 10\% while cost varied by at least 50x between the cheapest (Grok 4.1) and the most expensive (Gemini 3 Pro). For comparison, we estimated the cost for a 2-person LMIC expert panel in Africa to diagnose our cases to be \$2-\$40 per case. This is 4-80x more than the most expensive LLM and 200-4000x more than the cheapest LLM. Note that Gemini 3 Pro (Preview) used approximately $5\times$ more tokens per case due to its high-resolution default image settings. Exact token numbers used were not provided by Grok 4.1 and were estimated as an average of all other models, excluding Gemini-3 Pro (Preview) and Claude 4.1 Opus. }
    \label{fig:Post-calibration-Performance-vs-mean-USD-cost-per-case}
\end{figure*}

Since the costs per million tokens varied by $\sim 150\times$ across our 10 chosen models, it is natural to study the correlation between cost and LLM performance. Figure~\eqref{fig:Post-calibration-Performance-vs-mean-USD-cost-per-case}) shows the average score vs average cost per case, along with the cost-performance optimal Pareto frontier, which spans four models: Grok 4.1, GPT-5 Mini, Gemini 2.5 pro, GPT-5.1 from cheapest to most expensive. Claude 4.1 Opus was the worst model both in terms of $S_3$ and $S_4$ scores and in terms of cost per million tokens (see Table~\eqref{tab:model_card}).  

Conversely, it is remarkable that for $\sim$  \$0.02 per case, GPT-5 Mini provides diagnostic performance within $5\%$ of the best model and Grok 4.1 is within 13\% of the top model for a cost of $\sim$\$0.01, approximately 15x cheaper than the top model. 

Because published per-case costing data for LMIC expert adjudication panels are limited, we estimated the range of panel costs from first principles using observed review times in our study and salary-based clinician cost-per-minute assumptions from Nigeria and South Africa, which we take as indicative lower and upper salary ranges in LMIC settings \cite{bhekisisa2024remuneration_report,who2016workforce2030}. Our panel costs were approximately \$40 per case while public-sector physician salaries in Nigeria were reported to be approximately US\$1,200 per annum \cite{bhekisisa2024SA_Nigeria}, yielding a lower estimate of \$2 per case. Even with a lower bound of \$2 per case, the most expensive models are 4x cheaper, while the cheapest model is 200x less per case. For the top-end of the LMIC range, the cheapest model (Grok 4.1) is approximately 4000x cheaper. Cost differences of this magnitude suggest that, if safety and integration challenges can be addressed, LLM-based diagnostic support could become economically feasible and beneficial at scale in high-volume LMIC health systems where affordability and robustness are key constraints. Given the rapid decline in LLM costs over recent years, the affordability of high-quality diagnoses is likely to improve further. 

\subsection{Case-to-case model variance}

\begin{figure}
    \centering
    \includegraphics[width=0.95\linewidth]{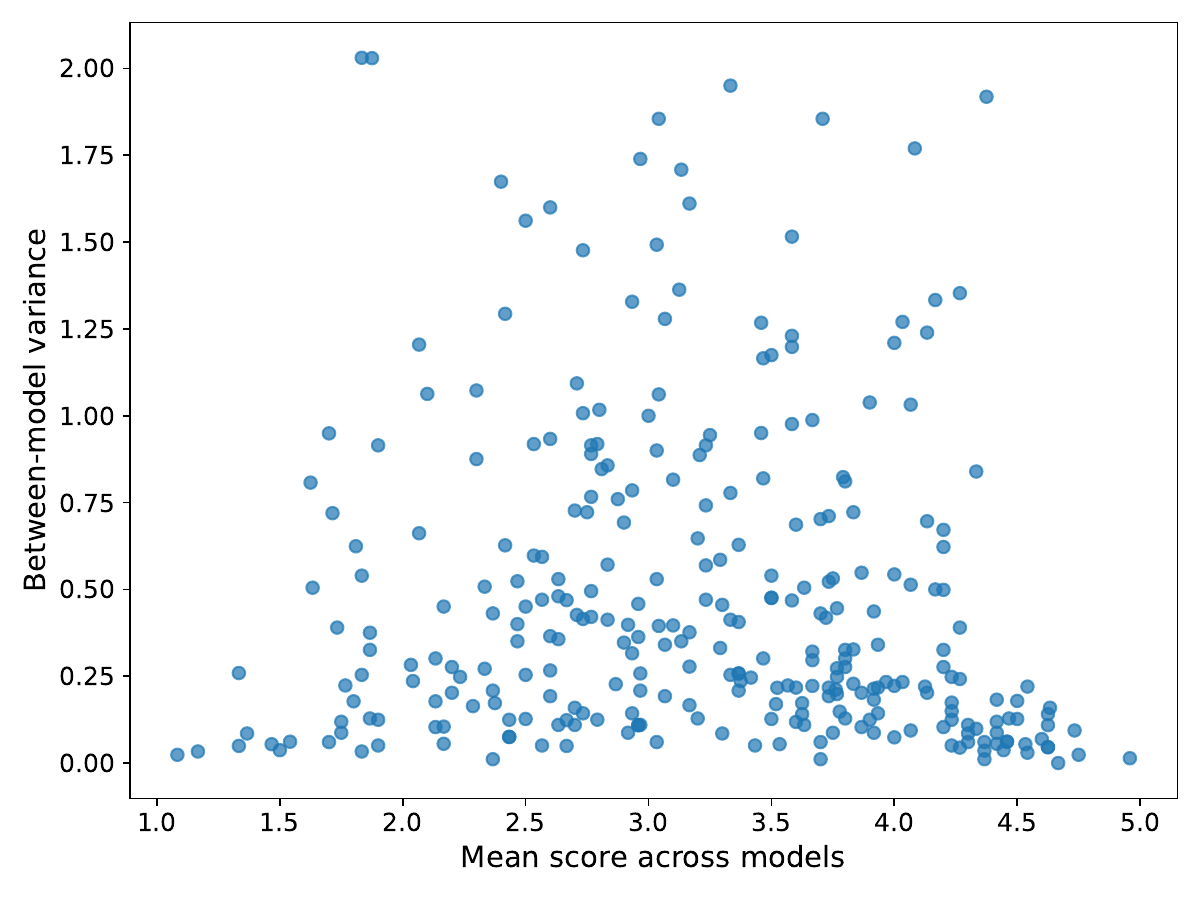}
    \caption{Model score variance (over the 10 LLMs) against mean model score for each of the 300 cases. Cases with low (high) mean score correspond to cases where all models disagreed (concurred) with the expert panel diagnosis. High between-model score variance corresponds to cases where there was significant disagreement between models about the true diagnosis. The tie-breaker cases were selected by choosing the cases with lowest mean model score (i.e. most of the models scored badly relative to the expert panel diagnosis). }
\label{fig:case_difficulty}
\end{figure}

Figure~\eqref{fig:case_difficulty} characterises case-level difficulty and model consensus by jointly analysing the mean performance of our ten LLMs against the expert panel and the variance of those scores across models. Cases with low mean scores represent scenarios in which models consistently fail to align with expert diagnoses, indicating systematically challenging or poorly specified clinical presentations. Conversely, high mean scores reflect cases where models reliably reproduce expert conclusions, suggesting clearer diagnostic signals or more standardised patterns. Importantly, variance across models disentangles agreement from correctness: high variance identifies cases where models diverge substantially in their predictions, pointing to ambiguity or sensitivity to model-specific inductive biases, while low variance indicates strong consensus—whether correct or incorrect. The tie-breaker cases discussed in Section~\eqref{sec:tie-breaker} were chosen from the low-mean, low-variance region, which isolates instances where all models disagreed with the panel, providing a targeted subset for deeper analysis of shared limitations of expert panels and current multimodal LLMs.

\subsection{Head-to-head win rates}

\begin{figure}
    \centering
    \includegraphics[width=0.99\linewidth]{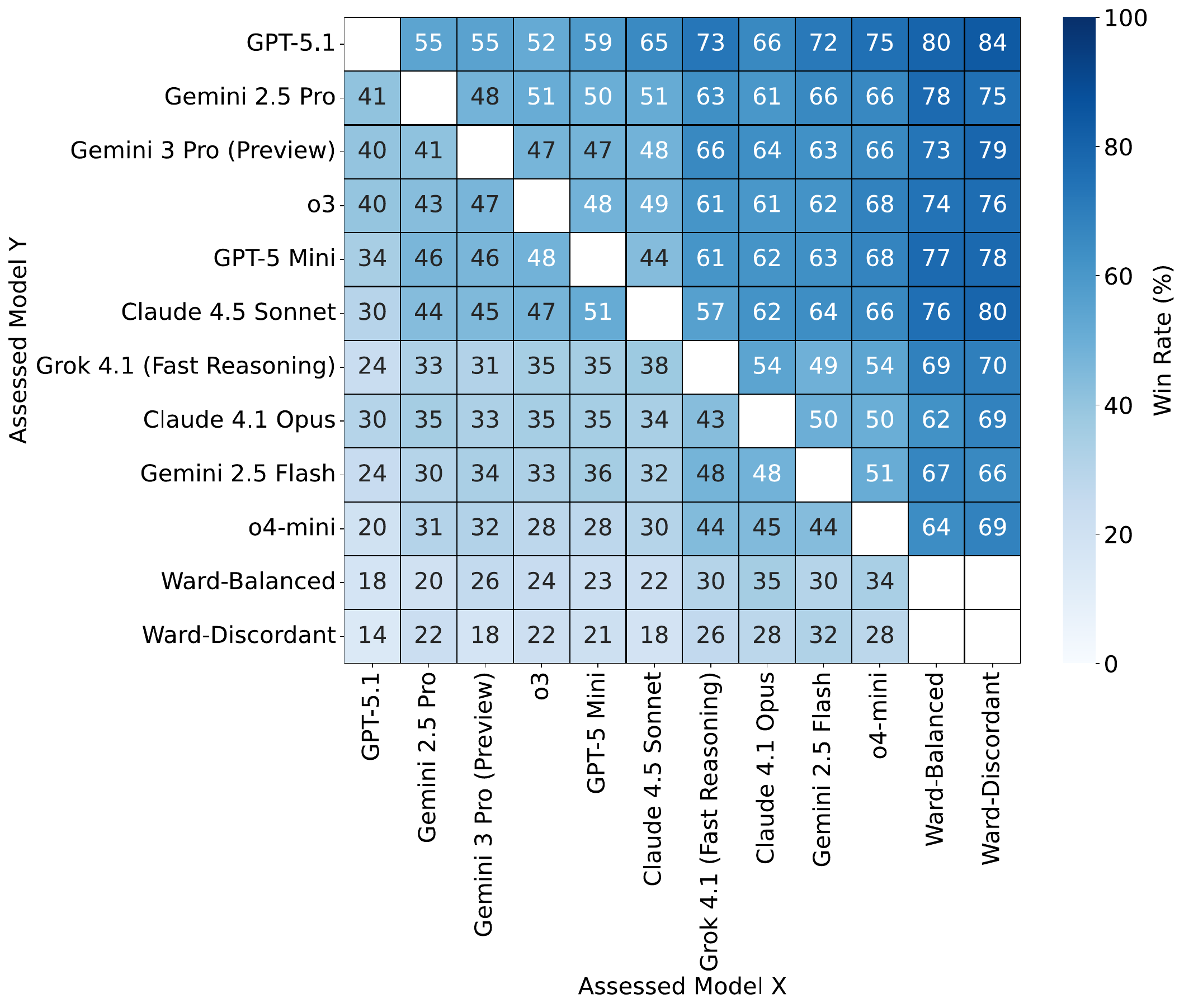}
    \caption{Win rates percentages for all models and ward cases using average raw $S_3$ scores on a case-by-case basis. We see that GPT-5.1 had a $>50\%$ win record against all other models, and $\geq 79\%$ win-rate against the ward.  All models had at least a 62\% win rate against ward diagnoses. Note that ties are not included, which were often around $3\%$ of cases.}
    \label{fig:Win-rates-percentages}
\end{figure}

Mean scores are a useful summary metric, but do not, of course, capture the full distributions. More insight comes from computing the head-to-head win rate for all the LLMs and against the ward diagnoses; see Figure~\eqref{fig:Win-rates-percentages}, which shows the percentage of head-to-head wins. The picture is consistent: GPT-5.1 has the best win rates ($\geq 52\%$ against all models) and the ward diagnoses have the worst, with the ward winning $\leq 18\%$ of cases against GPT-5.1.

\subsection{Correlations between models}

\begin{figure}
    \centering
    \includegraphics[width=0.99\linewidth]{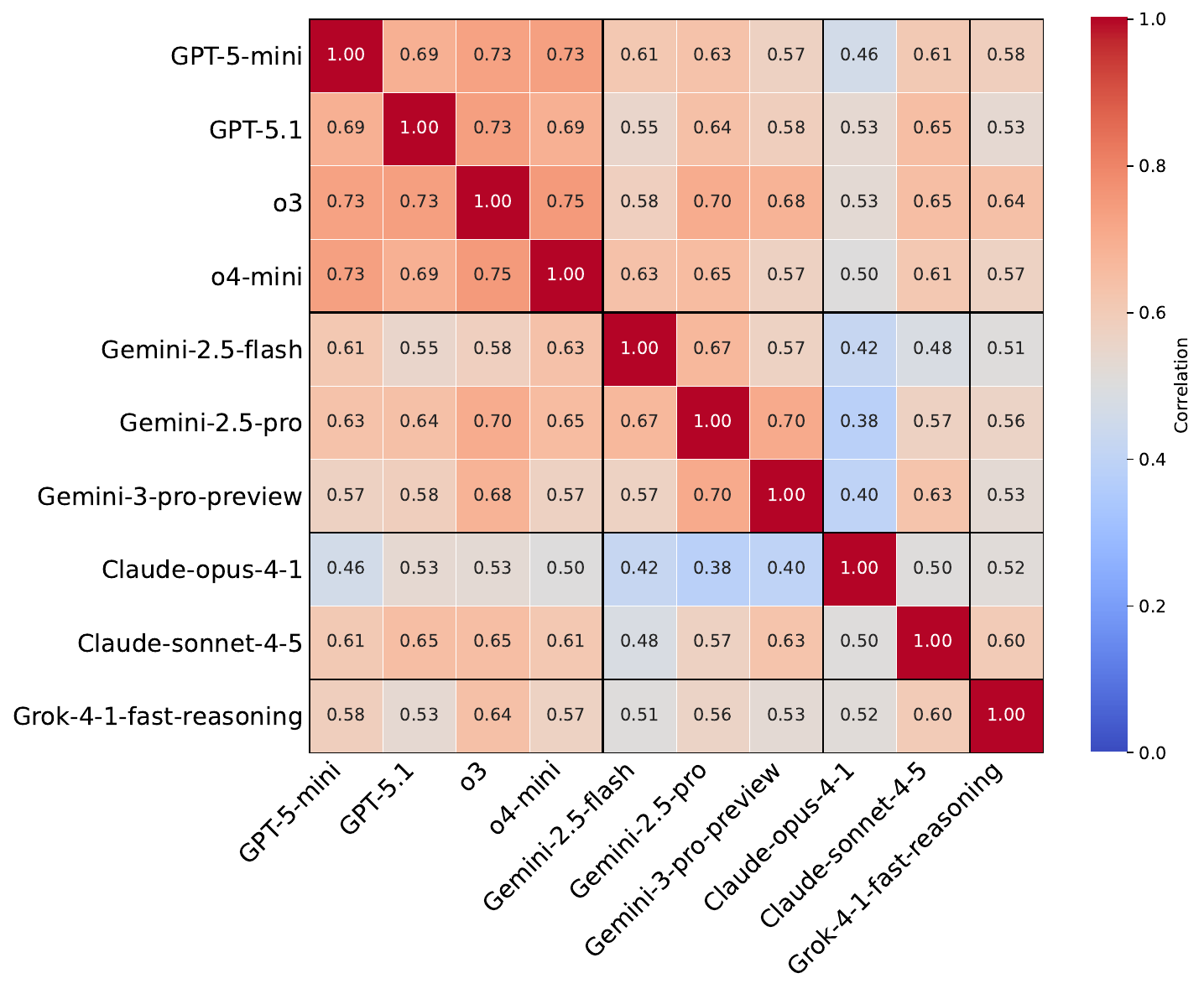}
    \caption{Spearman correlations between model $S_4$ scores across the 300 panel reviewed cases, grouped into vendor families (boundaries denoted by black lines). In general, we note high off-diagonal correlations ($\sim 0.59$) between models, suggesting common training data, reasoning or architecture design. The OpenAI models family is the most highly correlated family, while Claude 4.1 Opus is the maximally uncorrelated model, even when compared with Claude 4.5 Sonnet. } 
    \label{fig:model_corr}
\end{figure}

In Figure~\eqref{fig:model_corr}, we show pairwise Spearman correlations between the LLMs. This quantifies the extent to which different models exhibit similar case-level scores across the 300 panel cases. High correlations mean that two models tend to assign relatively higher and lower scores to the same cases, implying shared patterns of success and failure, while lower correlations indicate more distinct behaviour. Importantly, this does not by itself demonstrate anything about intrinsic case difficulty: shared low performance could reflect clinically meaningful challenges, but it could equally arise from shared sensitivities to artefacts, formatting, benchmark composition, or other nuisance features or, as discussed in Section~\eqref{sec:tie-breaker}, errors in expert panel diagnoses. It is clear that models are not behaving independently, and that some of this dependence appears to follow vendor-family boundaries. Stronger within-family concordance is consistent with the idea that models from the same lineage retain common behavioural signatures, plausibly reflecting overlap in training data, architectural design, multimodal integration strategy, and post-training objectives. At the same time, correlations well below 1 show that these shared influences are incomplete: the models are neither interchangeable nor simple replicas, and their residual differences imply meaningful model-specific variation in diagnostic behaviour.

The most significant example of this is provided by Claude 4.1 Opus, which is the most behaviourally distinct model in the set. Given that it also had the lowest mean score overall (see Table~\eqref{tab:all_scores}), this suggests that its distinct case-level scoring profile is more likely to reflect idiosyncratic or poorly calibrated behaviour than a beneficially different diagnostic perspective. In other words, Opus not only disagrees more often with the other models, but when it does so it also tends to disagree with the expert panel reference too.

\begin{table*}[htbp]
\centering

\renewcommand{\arraystretch}{1.2}
\begin{tabular}{p{0.13\linewidth} p{0.07\linewidth} p{0.07\linewidth} p{0.08\linewidth} p{0.07\linewidth} p{0.07\linewidth} p{0.07\linewidth} | p{0.07\linewidth} p{0.07\linewidth} p{0.03\linewidth}}
\hline
\multicolumn{1}{c}{} &
\multicolumn{6}{l}{\textbf{RAW LLM JURY SCORES}} &
\multicolumn{2}{l}{\textbf{PANEL-CALIBRATED SCORES}} &
\\
\cline{2-7} \cline{8-9}
\textbf{Assessed Model} & \textbf{Dx} & \textbf{DDx} & \textbf{Clinical Reasoning} & \textbf{Patient Safety} & 
\textbf{$\mathbf{S_4}$ (RAW)} &
\textbf{$\mathbf{S_3}$ (RAW)}&
\textbf{$\mathbf{S_4}$ (ISO)}&
\textbf{$\mathbf{S_3}$ (ISO)}&
\textbf{Cases} \\
\hline

GPT-5.1 & \textbf{3.36} & \textbf{3.04} & \textbf{3.49} & \textbf{3.50} & \textbf{3.41} & \textbf{3.35} & \textbf{4.04} & \textbf{4.01} & 288 \\

Gemini 2.5 Pro & 3.28 & 2.82 & 3.31 & 3.40 & 3.28 & 3.24 & 3.98 & 3.96 & {\bf 300} \\

Gemini 3 Pro (Preview) & 3.29 & 2.64 & 3.46 & 3.43 & 3.32 & 3.22 & 4.01 & 3.95 & 298 \\

GPT-5 Mini & 3.17 & 2.92 & 3.37 & 3.34 & 3.26 & 3.19 & 3.99 & 3.95 & 287 \\

o3 & 3.25 & 2.80 & 3.32 & 3.32 & 3.25 & 3.19 & 3.99 & 3.94 & 290 \\

Claude 4.5 Sonnet& 3.24 & 2.71 & 3.43 & 3.33 & 3.27 & 3.17 & 3.97 & 3.93 & 197 \\

Grok 4.1 (Fast Reasoning) & 2.98 & 2.56 & 3.15 & 3.06 & 3.01 & 2.93 & 3.85 & 3.80 & 288 \\

Gemini 2.5 Flash & 2.96 & 2.48 & 3.03 & 3.10 & 2.97 & 2.92 & 3.83 & 3.80 & {\bf 300} \\

o4-mini & 2.96 & 2.47 & 2.97 & 3.00 & 2.92 & 2.88 & 3.80 & 3.77 & 291 \\

Claude 4.1 Opus & 2.91 & 2.57 & 3.01 & 2.98 & 2.92 & 2.87 & 3.80 & 3.76 & 195 \\
\hline
Ward-Balanced & 2.88 & 1.99 & -- & 2.70 &--& 2.63 & -- & 3.65 & 139 \\

Ward-All & 2.72 & 1.99 & -- & 2.54 & -- & 2.50 & -- & 3.57 & 300 \\

Ward-Discordant & 2.58 & 1.99 & -- & 2.39 & -- & 2.39 & -- & 3.50 & 161 \\

\hline
\end{tabular}
\vspace{0.95em}
\caption{Mean model scores (on scale 1-5) averaged across all three LLM Jury models scores and across all cases where a response was returned.  The $S_4$ score averages the diagnosis score (Dx), differential diagnosis (DDx), clinical reasoning and patient safety scores with weights (0.3, 0.1, 0.3, 0.3) while "$S_3$" weights these scores (0.4, 0.2, 0, 0.4) to allow accurate comparison with Ward diagnoses. Columns to the left show raw LLM Jury averages, while columns to the right ($S_4$ (ISO) and $S_3$ (ISO)) show scores after isotonic regression of the LLM Jury scores to best fit the expert panel calibration scores. Note that only Gemini-2.5 Pro and Flash had perfect response rates, with Claude models averaging only a 65\% response rate due to the limits on the number of accepted input images (see final column). The impact of this exclusion changed final scores by 0.01. Bold indicates the assessed model with the best mean score and the assessed models that had a 100\% response rate.
}
 \label{tab:all_scores}
\end{table*}

Principal Component Analysis (PCA) provides a complementary view of the variation in case-level scores across models. Here, each model is represented by its vector of scores over the 300 benchmark cases, and PCA asks whether the differences between these model score profiles can be summarised by a few dominant underlying patterns. In this case, the first principal component explained only 19.3\% of the total variation in these model-by-case scoring patterns, with later components explaining similar amounts (6 PCs required to explain 80\% of variance). This indicates that differences between models are not mainly driven by a single shared axis. In practical terms, the models do not behave like near-identical systems with only small calibration differences. Instead, their similarities and differences are spread across several distinct dimensions of behaviour. Taken together, the correlation heatmap and PCA therefore suggest that frontier multimodal diagnostic models show substantial but incomplete concordance, including some family-level resemblance, while still occupying a moderately high-dimensional behavioural space shaped by multiple overlapping influences rather than one dominant consensus pattern.

\subsection{Serious safety risk rate}
\label{sec:safety}

Focussing on average scores and even win rates carries with it the dangers of missing the distributions of scores; the sequences \{4,4,6,6\} and \{1,1,9,9\}
have the same mean but are clearly very different. One area where this is particularly important is high-risk diagnoses. Despite all models having a higher average $S_3$ score than the ward, is it possible that some models have higher serious risks to patient safety? 

To test this, we computed, for each model, the percentage of cases whose patient safety score was below a threshold $T$. For the RAW LLM Jury scores we used $T = 2$, as scores of 1 and 2 indicate a major or critical risk to patient safety. In our prompt, we instructed the LLM Jury that a safety score of 1 or 2 corresponded respectively to: 

{\bf Critical harm likely}: Wrong treatment could be fatal or cause severe permanent morbidity; or critical delay in life-saving therapy.

{\bf Major harm probable}:  Significant morbidity, organ damage, or dangerous delay likely in therapy.

For the isotonic panel-calibrated, we chose $T = 2.5$, which, given that the calibrated scores are approximately 1 unit higher than the raw scores, effectively corresponds to a significantly more stringent threshold akin to 1.5 in the raw scores. We refer to these as severe safety risk cases \footnote{Since these scores are computed using the expert panel diagnoses as ground truth and the expert panels are not perfect, a low safety score does not necessarily imply a true safety risk to the patient. }

Table~\eqref{tab:safety_tail} shows that the ward diagnoses contain the highest fraction of scores below $2$, totalling 22\% of patient records. The top six models all have severe safety risk rates $\leq 11\%$, half the rate of the ward. In the case of the calibrated scores with $T = 2.5$, both ward and all models have severe risk rates of $< 3\%$. Although the models appear to outperform the ward in terms of severe risk, none were free from concerning safety scores, underscoring the need for governance mechanisms even when average safety performance is acceptable. We end, however, by pointing the reader to Section~\eqref{sec:tie-breaker}, which adds significant nuance to the identification of severe safety risk. 

\begin{table*}[htbp]
\centering
\renewcommand{\arraystretch}{1.2}
\begin{tabular}{p{0.16\linewidth} p{0.06\linewidth} p{0.07\linewidth} p{0.1\linewidth} 
| p{0.07\linewidth} p{0.1\linewidth}}
\hline
& & \multicolumn{2}{c}{\textbf{Raw Mean LLM Jury Scores ($\mathbf{T = 2}$)}} & \multicolumn{2}{c}{\textbf{Mean Calibrated Scores ($\mathbf{T = 2.5}$)}} \\
\cmidrule(lr){3-4} \cmidrule(lr){5-6}
\textbf{Model Name} & $\mathbf{n_{\text{\textbf{total}}}}$ & $\mathbf{n < T}$ & \textbf{Fraction $\mathbf{< T}$} & $\mathbf{n < T}$ & \textbf{Fraction $\mathbf{< T}$} \\
\hline

Gemini 2.5 Pro & 300 & 29 & 0.10 & 4 & 0.01 \\

GPT-5 Mini & 287 & 28 & 0.10 & 0 & 0 \\

GPT-5.1 & 288 & 29 & 0.10 & 1 & 0 \\

Gemini 3 Pro (Preview) & 298 & 33 & 0.11 & 3 & 0.01 \\

Claude 4.5 Sonnet  & 197 & 22 & 0.11 & 0 & 0 \\

o3 & 290 & 33 & 0.11 & 3 & 0.01 \\

Gemini 2.5 Flash & 300 & 45 & 0.15 & 4 & 0.01 \\

Grok 4.1 (Fast Reasoning) & 288 & 47 & 0.16 & 4 & 0.01 \\

o4-mini & 291 & 51 & 0.18 & 5 & 0.02 \\

claude 4.1 Opus  & 195 & 35 & 0.18 & 4 & 0.02 \\
\hline
Ward-Balanced & 139 & 28 & 0.20 & 1 & 0.01 \\

Ward-Discordant & 161 & 37 & 0.23 & 4 & 0.02 \\

\hline
\end{tabular}
\vspace{0.95em}
\caption{Safety summary for model and ward diagnoses. We show the fraction of cases where the mean $S_3$ score was below a key threshold ($T$). The left half of the table (“Raw Mean LLM Jury Scores”) shows results with threshold $T = 2$ using the raw LLM Jury scores. The right shows results with $T = 2.5$ using the expert-panel-calibrated LLM Jury scores. In both cases, ward diagnoses had the worst fraction of low-scoring cases (20\% and 23\% for the Balanced and Discordant case subsets,  respectively). }
\label{tab:safety_tail}
\end{table*}

\subsection{Can models reason clinically?}

Given the excellent diagnostic performance of LLMs we have presented, one interesting question is whether the LLMs are reasoning like physicians or whether they are simply performing highly successful, but superficial, pattern matching. This question has added nuance given our choice of LLMs, which mixes high-end "reasoning" models (like GPT-5.1, Gemini 3 Pro and 4.1 Opus) and cheap "non-reasoning" models. 

Truly addressing this question is deep and beyond the scope of this work. We can, however, probe some aspects of this question through the clinical reasoning responses we demanded from all of the models to justify their combined diagnoses.  Two examples are provided in supplementary appendix~\ref{sec:clinical_reasoning_examples}. These LLM reasonings must reference the relevant patient data and closely match the expert panel clinical reasoning to score high.

\begin{figure}
    \centering
    \includegraphics[width=0.9\linewidth]{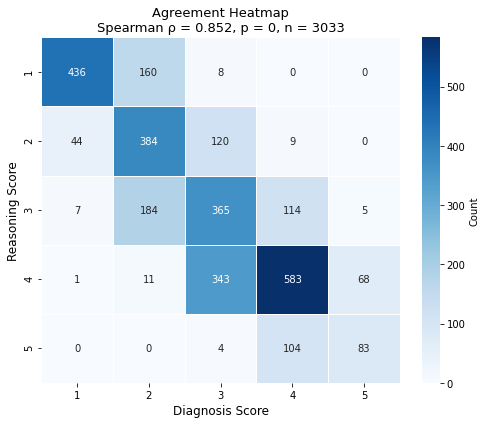}
    \caption{Distribution of LLM diagnosis and clinical reasoning scores across all 3033 case-model combinations. The Spearman $\rho$ correlation is high (0.85) and the near-diagonal concentration of the joint distribution, together with the restriction of disagreements to adjacent categories, indicates that diagnosis and reasoning outputs are tightly coupled at the case level. This is inconsistent with models with generic clinical reasoning or superficial diagnostics. }
    \label{fig:reasoning}
\end{figure}

Figure~\eqref{fig:reasoning} shows binned data over all 10 LLMs for model clinical reasoning scores and model diagnosis (Dx) scores. We see that there is a high correlation between them (Spearman's $\rho = 0.85$). In particular, across all models, there were no examples where the Dx and clinical reasoning scores differed by 3 or more points. There is substantial dependence: the mutual information between them was 0.90 bits, corresponding to approximately 41\% reduction in uncertainty in either variable given the other. This implies that there is a very tight connection between the diagnoses and the clinical reasoning in all tested models, and rules out superficial pattern-matching diagnosis.

\subsection{Tie-breaker results}
\label{sec:tie-breaker-results}

As discussed in Section~\eqref{sec:tie-breaker}, our most senior physicians took part in tie-breaker panels that re-diagnosed 20 cases selected as the most Discordant between models and expert panel diagnoses. 

\begin{figure}
    \centering
    \includegraphics[width=\linewidth]{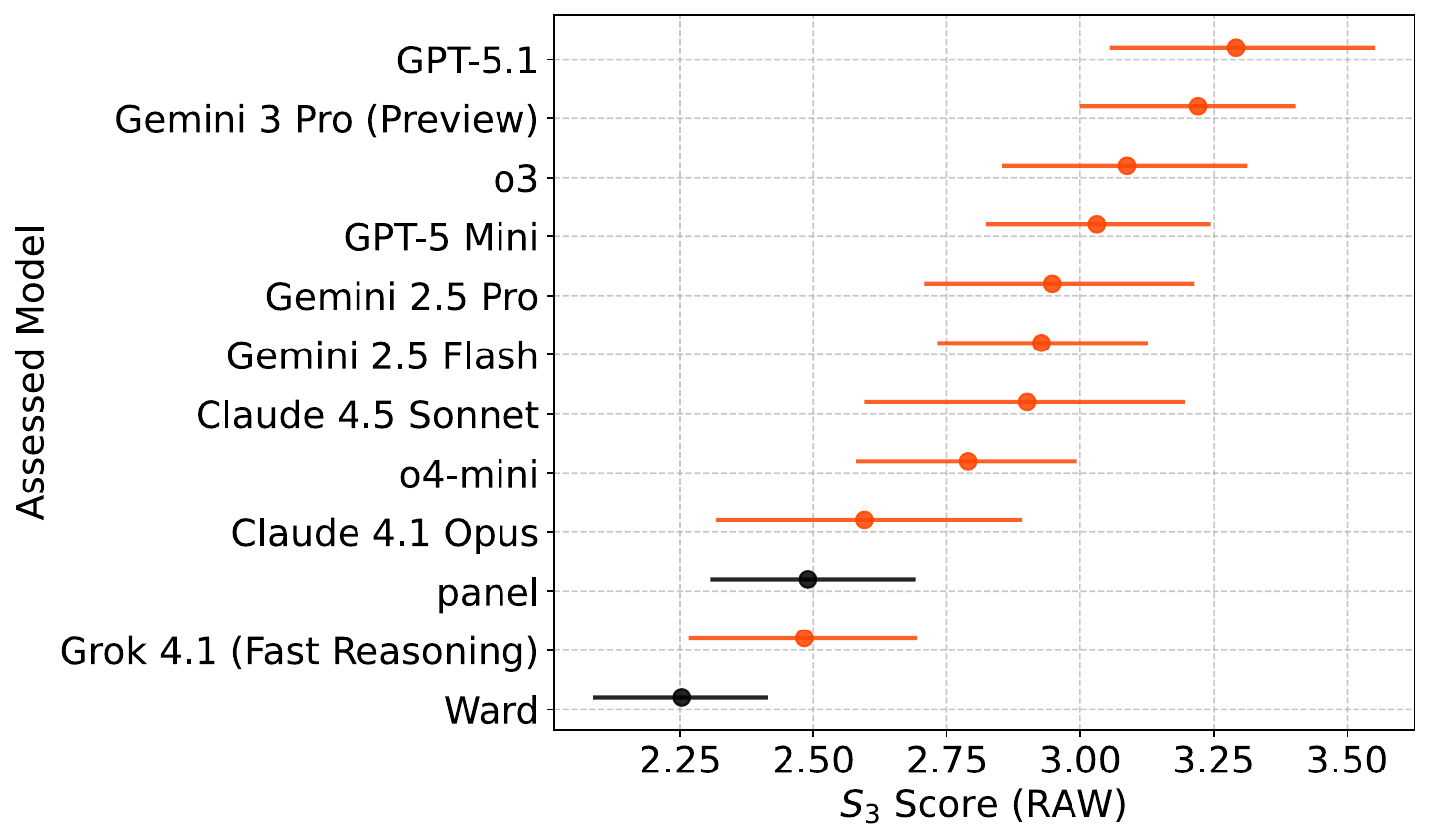}
    \caption{{\bf Tie-breaker performance} of AI models vs expert panels and ward using RAW LLM Jury scores with the most senior tie-breaker panel diagnoses as ground truth. 68\% confidence intervals on means are also shown. We see that the best models are significantly better than even the expert panels and only Grok 4.1 had a mean score below the expert panel. While the top models had very little drop in score relative to the full 300 case dataset, the ward score dropped 0.25 points (approximately a 10\% drop). }
    \label{fig:tie-breaker}
\end{figure}

Using the tie-breaker diagnoses as the new ground truth, we were able to score models, ward and expert panels on an equal footing. 
Figure~\eqref{fig:tie-breaker} shows the results and makes an uncomfortable point: while in only 20 cases the senior tie-breaker shifts the centre of gravity toward the models, not the expert panels. GPT-5.1 still ranks clearly first, with Gemini 3 Pro (Preview) second. But 9 of the 10 LLMs outperform the expert panels, which in turn outperform both Grok 4.1 and the Ward. 

This implies that a meaningful fraction of the supposed LLM “failures” discussed in Section~\eqref{sec:safety} may well have been failures of the expert panel adjudications, despite the precautions taken in selection of panel members. That is a stronger conclusion than simple label noise: it suggests that, in the hardest, most disputed cases, frontier models may have been closer to the best available diagnosis than the expert panels used to judge them. The implication is consequential—benchmarking clinical AI against a single panel diagnosis can systematically understate model performance precisely where diagnostic uncertainty is greatest.

Table~\eqref{tab:tie-breaker} provides additional detail with estimated 95\% confidence intervals on the means. The final column reports the probability that each LLM outperforms the expert panels on this subset; for two models—GPT-5.1 and Gemini 3 Pro (Preview)—this probability exceeds 95\%.

\begin{table*}[htbp]
\centering

\renewcommand{\arraystretch}{1.2}
\begin{tabular}{p{0.18\linewidth} p{0.16\linewidth} p{0.12\linewidth} p{0.12\linewidth} p{0.12\linewidth} p{0.16\linewidth}}
\hline
\textbf{Model Name} & \textbf{$\mathbf{S_3}$ Score Mean} & \textbf{Change in Score} & \textbf{Lower 95\% CI} & \textbf{Upper 95\% CI} & \textbf{P(Model > Expert Panel)} \\
\hline

GPT-5.1 & 3.29 & -0.06 & 3.01 & 3.61 & 97.6\% \\

Gemini 3 Pro (Preview) & 3.22 & 0.00 & 2.99 & 3.47 & 97.4\% \\

o3 & 3.09 & -0.10 & 2.80 & 3.38 & 90.0\% \\

GPT-5 Mini & 3.03 & -0.16 & 2.75 & 3.33 & 88.6\% \\

Gemini 2.5 Flash & 2.97 & +0.05 & 2.70 & 3.24 & 87.7\% \\

Gemini 2.5 Pro & 2.95 & -0.29 & 2.65 & 3.26 & 86.4\% \\

Claude 4.5 Sonnet & 2.96 & -0.21 & 2.56 & 3.28 & 74.7\% \\

o4-mini & 2.79 & -0.09 & 2.53 & 3.05 & 74.3\% \\

Claude 4.1 Opus & 2.60 & -0.27 & 2.27 & 2.92 & 45.6\% \\

\hline
{\bf Expert panel} & {\bf 2.57} & {\bf --} & {\bf 2.33} & {\bf 2.79} & {\bf --} \\
\hline

Grok 4.1 (Fast Reasoning) & 2.48 & -0.45 & 2.26 & 2.74 & 39.8\% \\

Ward & 2.25 & -0.25 & 2.07 & 2.45 & 15.0\% \\

\hline
\end{tabular}
\vspace{0.95em}
\caption{Performance of models and expert panels/ward using the diagnoses from the tie-breaker panels formed by the most senior clinicians as ground truth. We see minimal drop in the scores of the top models relative to the full sample of 300 cases (c.f. Table~\eqref{tab:all_scores}). Interestingly, the expert panel score (2.57) is lower than 7 of the 10 AI models, though higher than the ward score (2.25). In the last column, we show the probability (estimated from 1000 bootstrap samples) that each model is actually better than the expert panels. Both GPT-5.1 and Gemini 3 Pro (Preview) are better than the expert panels at greater than 95\% confidence.  This must be considered only indicatively because the 20 cases were not randomly chosen but rather were chosen because of the high level of discordance between model and expert panel diagnoses, but the superior performance of the models over the panels provides strong additional evidence for the quality of the best AI models. The AI models and ward dropped 0.14 and 0.27 points, respectively, compared to the full sample (see Table~\eqref{tab:all_scores}). The AI model mean was 2.93, 0.36 points higher than the expert panel. }
\label{tab:tie-breaker}
\end{table*}

\subsection{Impact of radiological reports on performance}

Analysis of the tie-breaker data discussed in Section~\eqref{sec:tie-breaker} revealed that the LLMs had been provided with only 128 x 128 pixel low-resolution images of all radiological reports, making them completely unreadable to the human eye. As a result, all cases had to be rerun with high-resolution images (512 pixel horizontal resolution with vertical resolution set by the aspect ratio of the pdf). It is these latter results that we have reported in the rest of this paper. However, this error provided a natural and unintended ablation test of the importance of the radiological report to the LLM diagnosis. 

As shown in Table~\eqref{tab:mean_combined_score_highres_lowres_delta},  replacing high-resolution radiology report images with unreadable low-resolution radiology report images led to a consistent reduction in mean $S_4$ score for most assessed models. In all cases, the models received the actual radiology imaging data. This effectively isolates the incremental value of access to readable radiology report content in a multimodal diagnostic setting. 

The overall pattern suggests that, for most models, diagnostic performance benefits materially from the textual information contained in radiology reports, even when the underlying imaging is also provided. The only exception was Claude 4.1 Opus, whose performance was unchanged. Limited experimentation showed that models provided with only the low-resolution reports and no imaging did sometimes hallucinate diagnoses. Future work should explore both the anomalous Claude 4.1 Opus result and whether the low-resolution radiology reports had any impact compared to providing no report at all: were the low-resolution reports simply ignored or did they regularly induce hallucinations?

\begin{table}[htbp]
\centering
\renewcommand{\arraystretch}{1.2}
\begin{tabular}{p{0.32\linewidth} | p{0.13\linewidth} p{0.13\linewidth} p{0.12\linewidth}}
\hline
& \multicolumn{3}{c}{\textbf{{\bf Mean} $\mathbf{S_4}$ {\bf Score}}} \\
\cmidrule(lr){2-4}
\textbf{Assessed Model} & \textbf{High-Res} & \textbf{Low-Res} & \textbf{$\mathbf{\Delta}$} \\
\hline
Gemini 2.5 Pro & 3.28 & 2.90 & 0.37 \\
Claude 4.5 Sonnet & 3.26 & 3.17 & 0.09 \\
GPT-5 Mini & 3.26 & 3.05 & 0.21 \\
o3 & 3.25 & 3.10 & 0.15 \\
Gemini 2.5 Flash & 2.98 & 2.82 & 0.16 \\
o4-mini & 2.92 & 2.78 & 0.15 \\
Claude 4.1 Opus & 2.90 & 2.91 & -0.01\\
\bottomrule
\end{tabular}
\vspace{1em}
\caption{Mean $S_4$ scores per assessed model in the high and low resolution (High-Res and Low-Res) radiology report conditions. $\Delta$ is defined as High-Res minus Low-Res. While most models scored significantly lower in the low-resolution condition, Claude 4.1 Opus was the only model unaffected. }
\label{tab:mean_combined_score_highres_lowres_delta}
\end{table}

\subsection{Impact of the number of images on performance}

The fact that the Anthropic models only returned diagnoses for the shortest 65\% of cases leads to a question of whether this gave them an unfair advantage. To test the impact of this on results, and implicitly the impact of the number of images (noting that all MRIs and CT-scans are converted to ordered images), we computed the performance of all models on the subset of 195 cases for which all models returned a diagnosis, see Table~\eqref{tab:sub3}. 

Comparing the results shown in Table~\eqref{tab:sub3} and  Table~\eqref{tab:all_scores}, we see very little difference. GPT-5.1 and the large Gemini models (2.5 and 3) still occupy the top three spots and all models and the ward have similar scores. From this, we can conclude that using the full data provides a fair comparison of models and that cases with many images, typically indicating more than one MRI or CT scan, did not prove more difficult to diagnose. This may be related to the availability of the radiology reports, a hypothesis to test in future work. 

\section{Discussion}

The VALID datasets presented here were designed as an interlocking benchmark enabling simultaneous evaluation of diagnostic accuracy, adjudication dynamics, information sensitivity, and evaluator reproducibility across human and AI decision-makers.

The core dataset is formed by 300 multimodal inpatient cases from a large South African public-sector tertiary hospital evaluated by two-person expert panels for combined primary and secondary diagnoses, differential diagnosis and clinical reasoning. Using this as ground truth, we evaluate both routine ward diagnoses and both diagnoses and clinical reasoning of ten leading multimodal LLMs. We briefly discuss here the main findings. 

\subsection{Cost-Performance Tradeoff}

\begin{figure}
    \centering
    \includegraphics[width=\linewidth]{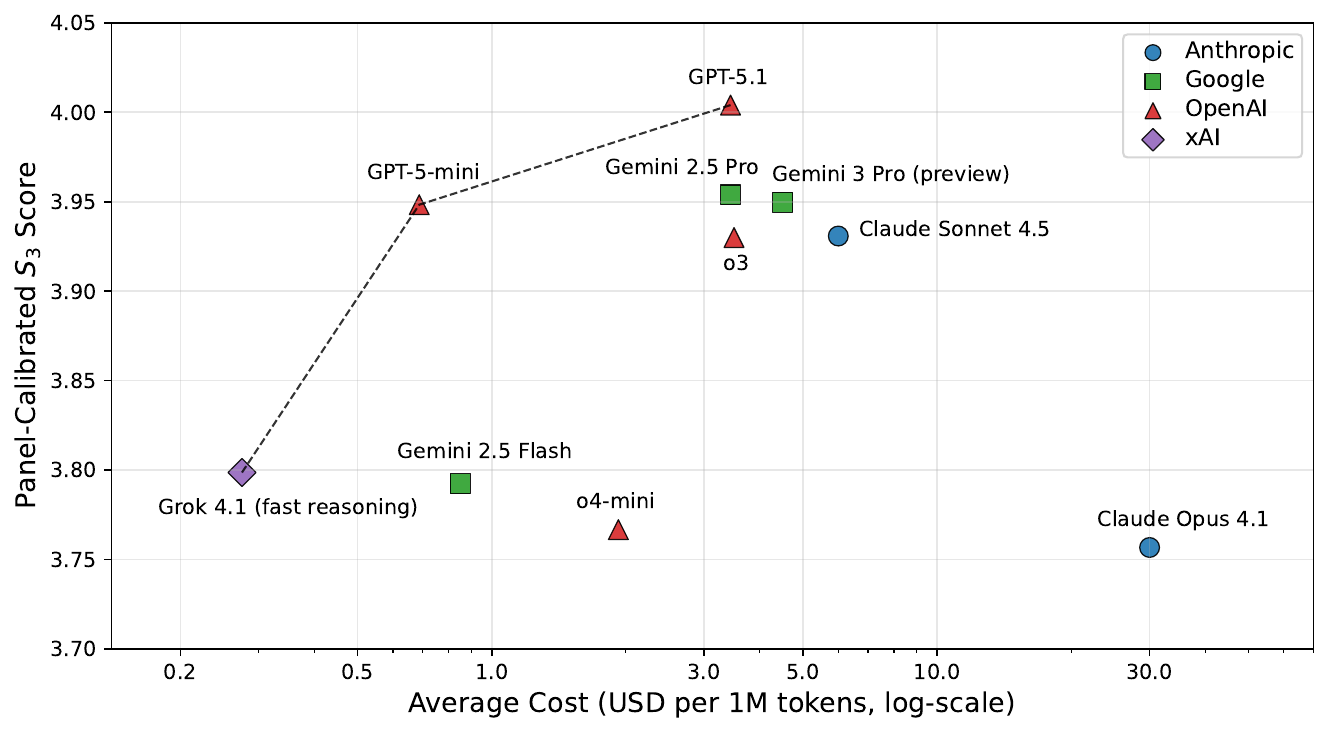}
    \caption{{\bf Mean $S_3$ LLM scores after calibration vs per-million-token cost}.  The Pareto frontier (dashed line) shows optimal performance as a function of model cost (average of input + output costs in USD per million tokens). Performance between models varied little while cost varied by $150\times$ between the cheapest (Grok-4.1) and the most expensive (Claude Opus 4.1). Notice that while Gemini 3 Pro preview costs look similar to those for Gemini 2.5 Pro, its image tokeniser uses significantly more tokens than other models, raising the cost per case significantly; c.f. Figure~\eqref{fig:Post-calibration-Performance-vs-mean-USD-cost-per-case}, which shows performance vs cost per case.}
    \label{fig:Post-calibration-Performance-vs-token cost}
\end{figure}

All models demonstrated remarkably similar average diagnostic performance, despite 50-fold variation in cost per case (see Figure~\eqref{fig:Post-calibration-Performance-vs-token cost}). Across models, differences in composite scores $S_3$ and $S_4$ were small, confidence intervals often overlapped, and model rankings were sensitive to minor shifts in means. These findings suggest that among contemporary frontier-level multimodal models, mean diagnostic performance may be converging, even when underlying architectures, costs, and input-handling constraints differ substantially. 

\subsection{Comparison with Routine Ward Diagnoses}

A distinctive strength of this study is its use of routine ward diagnoses as a real-world comparator. Ward diagnoses reflect real conditions of practice in high-volume public-sector wards, where time pressure, incomplete information flow, and documentation constraints shape clinical decision-making. Against this backdrop, the LLMs significantly outperformed routine ward diagnoses across all average metrics (at 95\% confidence interval,  see Table~\eqref{tab:all_scores}. This superiority was also seen in the case-by-case win rate with the best models winning more than 75\% of cases against the ward. This comparison does not imply underperformance by clinicians but illustrates the baseline against which clinical AI tools must ultimately be evaluated: the noisy, resource-constrained realities of care delivery rather than idealised benchmark tasks.

\subsection{Safety Considerations and Tail-Risk}
Despite generally high mean safety scores, all models generated 10-18\% low-safety outputs (safety scores $\leq 2$), highlighting that tail-risk remains an important safety challenge for clinical deployment. We do emphasise that these numbers were lower than the ward diagnoses and are also likely affected by any erroneous ground truth diagnoses (as evidenced by the Tie-Breaker panels). However, even infrequent unsafe recommendations may have significant consequences in acute inpatient settings. The presence of safety-tail outputs across all models—including the highest performing—supports the argument that focusing solely on average diagnostic metrics is insufficient. Effective governance frameworks should therefore emphasise:
\begin{itemize}
    \item Real-time uncertainty signalling,
    \item Escalation pathways for ambiguous or high-risk cases, and
    \item Post-deployment monitoring that tracks safety-tail frequency. These mechanisms may be more consequential for patient outcomes than marginal differences in mean performance. 
\end{itemize}

\subsection{Implications for LMIC Public-Sector Hospitals}
For LMIC health systems—where specialist availability, investigation access, and documentation completeness vary—our findings carry several implications. First, the tight clustering of mean performance suggests that affordability, robustness under complex multimodal input loads, and input-limit constraints may be as important as absolute diagnostic performance in determining which models should be selected for deployment. Second, the fact that costlier, larger models did not consistently outperform lower-cost alternatives suggests that value-for-money analyses should accompany technical evaluations in resource-stretched settings. Third, the inclusion of real routine ward diagnoses grounds these findings in actual public-sector workflow conditions, which may differ substantially from environments where most LLM evaluations are conducted. 

An important example of a pragmatic factor is the large variation in the rate at which LLMs returned a diagnosis, which varied between 65\% and 99\%. This was primarily driven in this dataset by model limits on the maximum number of images (see Table~\eqref{tab:model_card}). Although this constraint is likely to drop in future, it is an example of a critical feature to monitor in real deployments.

\subsection{Strengths of the Study}

This study has several strengths: 
\begin{itemize}
    \item It uses rich, multimodal inpatient LMIC data that mirror the complexity and uncertainty of real admissions rather than synthetic or single-modality cases.
    \item Expert adjudication provides a rigorous reference standard for diagnosis, differential, and clinical reasoning.
    \item The inclusion of routine ward diagnoses as a comparison offers a pragmatic baseline rarely examined in LLM diagnostic evaluations.
    \item The use of a panel-calibrated LLM Jury allowed consistent, scalable evaluations of all metrics and all models on all cases while maintaining alignment with expert scoring. 
    \item The use of Rescore and Tie-Breaker panels allowed tests of inter-rater agreement, LLM Jury performance and insights into the origin of disagreements between expert panels and the ensembled LLM diagnoses.
    \item The inclusion of a free-text clinical reasoning as one of our metrics for understanding LLM performance provides insights beyond diagnoses. 
\end{itemize}   

\subsection{Limitations}

Several limitations warrant consideration. First, the scoring relied primarily on the LLM Jury after calibration rather than full expert scoring of the 10,000+ model outputs; although agreement was good and the Jury was consistently stricter than human panels, residual measurement bias cannot be excluded. 

Second, the cohort derives from a single large tertiary hospital. External validation across other LMIC settings, hospital levels, or case‑mix distributions was not feasible due to data‑access constraints but will be required to establish true generalisability.  Third, model non-response—driven largely by vendor-imposed image-limits—resulted in missing outputs for some models, particularly Claude models, which returned results for only $\sim 65\%$ of cases; this missingness reflects real input-constraint limitations of current multimodal systems. Fourth, models were evaluated in a single-pass, closed-book setting using standardised prompting; prompts and context engineering optimised on a per-model basis or iterative reasoning might yield different results. 

\subsection{Conclusions}

Taken together, our findings suggest that multimodal LLMs have reached  clinical usefulness not because a single model is overwhelmingly superior, but because several leading systems now deliver consistently strong diagnostic support on difficult, real-world multimodal inpatient cases from a resource-constrained public hospital setting. The broader implication is that, for LMIC health systems, the central question may shift from whether these models can add value to how they can be deployed safely, affordably, and at scale within existing care pathways. 

When relatively low-cost models perform close to the frontier, practical considerations such as price, context-window limitations, multimodal input handling, reliability, and governance become at least as important as small gains in headline accuracy. At the same time, the persistence of occasional unsafe recommendations underscores that LLMs should be implemented as supervised decision-support tools rather than autonomous diagnosticians until these issues are resolved. Overall, this study provides strong evidence that multimodal LLMs could become a highly cost-effective force multiplier for inpatient diagnostic care in under-resourced hospitals, provided deployment is paired with robust oversight, workflow integration, and safeguards against tail-risk errors.

\section*{Funding}
This study was funded by The Gates Foundation. While the funder provided useful input to aspects of the study design, they had no authority over the study design; data collection, management, analysis, interpretation, preparation, review, or approval of the manuscript; or the decision to submit for publication. 

\section*{Acknowledgements}
We thank Scott Mahoney and Jess Rees for valuable discussions and/or comments on the draft and Lungile Gabuza, Victor Mngomezulu, Merika Tsitsi, 
Sithembiso Velaphi for assistance in conducting the study and Ekram Asefa for software development contributions to  the panel platform.  

We thank the physicians who participated in the many expert panels: Anu Abraham, Shabbir Alekar, Constance Adams, Nirvana Bharuthram, Tasneem Bux, Dixit Dullabh, Lara Greenstein, Peter Hewsen, Ibraaheem Ismail, Ismail Kalla, Sanjay Lala, Anees Laher, Shannon Leahy, Binu Luke, Siliziwe Lusu, Nhlakanipho Mangeni, Jaishil Manga, Gugulethu Mapurisa, Farzahna Mahomed, David Moore, Pramone Moodley, Aqeela Moosa, Vivendra Naidoo, Sadiya Nanabhay, Bavinash Pillay, Brent Prim, Gary Reubenson, Bianca Rowe, Kyle Schnaar, Sharise Singh, Kebashni Thandrayen, Ismail Tickley and Nicole Van Wyk. 

We acknowledge Lungile Taye, Mahtaab Khan, Tamara Romanini, Christopher Stavrou and Cameron Fisher for assistance in case abstraction.

\printbibliography

\begin{table*}[htbp]
\centering
\renewcommand{\arraystretch}{1.4}
\begin{tabular}{p{0.17\linewidth} p{0.4\linewidth} p{0.07\linewidth} p{0.12\linewidth} p{0.12\linewidth}}
\hline

\textbf{Scored Dimension} & \textbf{Definition} & \textbf{Scale} & \textbf{$\mathbf{S_4}$ Score Weights} & \textbf{$\mathbf{S_3}$ Score Weights} \\
\hline

Diagnostic Accuracy & Agreement of combined primary and secondary diagnoses with expert reference standard. & 1--5 & 0.30 & 0.40 \\

Differential Diagnosis & Differential at admission. Scores match to expert panel reference, penalising omissions and extra differentials. & 1--5 & 0.10 & 0.20 \\

Clinical Reasoning Quality & Extent to which causal story and key discriminators used match the expert panel’s clinical reasoning. & 1--5 & 0.30 & 0 \\

Patient-Safety Score & Expected safety if management followed the model’s/ward’s diagnosis instead of the expert diagnosis. & 1--5 & 0.30 & 0.40 \\

\hline
\end{tabular}

\vspace{0.95em}

\caption{Scoring Dimensions and Weights. Ward diagnoses lacked explicit reasoning, so the composite weights were renormalised to 0.4 / 0.2 / 0 / 0.4 to allow fair comparison between LLMs and routine care.}
\label{tab:weights}
\end{table*}

\begin{figure}
    \centering
    \begin{subfigure}[t]{\linewidth}
        \centering
        \includegraphics[width=\linewidth]{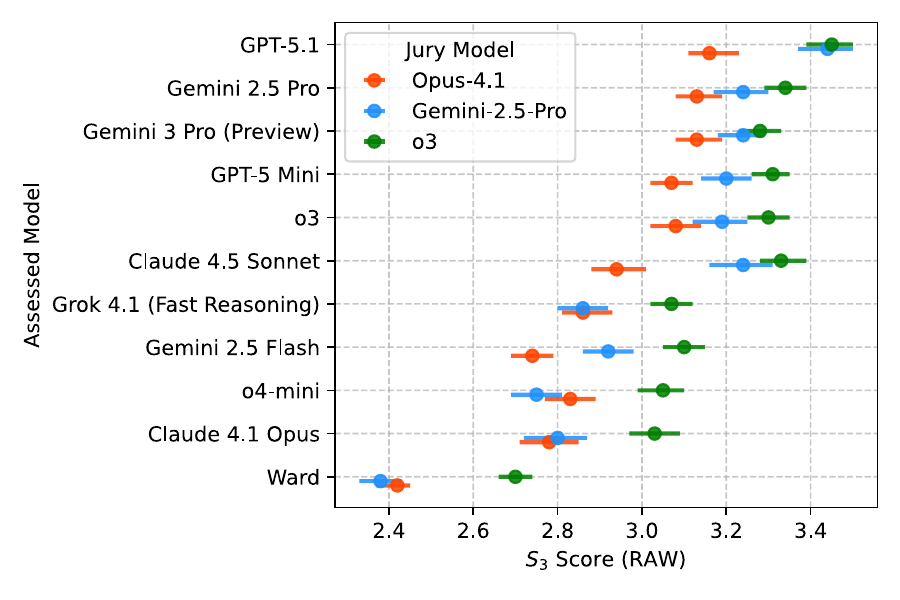}
        \caption{}
        \label{fig:scores-raw}
    \end{subfigure}
    \\
    \begin{subfigure}[t]{\linewidth}
        \centering
        \includegraphics[width=\linewidth]{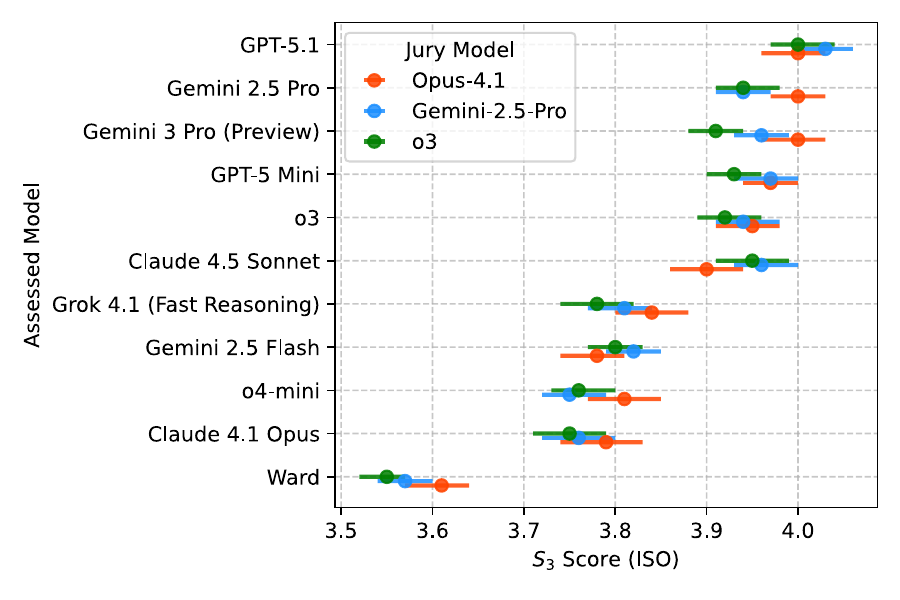}
        \caption{}
        \label{fig:score-cal}
    \end{subfigure}
    \caption{Mean $S_3$ scores from the three LLM Jury models (Gemini-2.5, o3, Opus-4.1) for all LLMs and the ward {\bf before} (a) and {\bf after} (b)  calibration of the LLM Jury scores onto the expert panel scores using per-jury isotonic regression. Error bars show 68\% confidence intervals and are slightly offset vertically for clarity. Before calibration o3 is the most lenient scorer and Opus-4.1 the strictest. The LLM Jury scores after calibration show good coherence for each tested model and good agreement with expert panels (not shown; see \cite{Rouillard2026LLMJury}). }
    \label{fig:Universal-Weighted-Scores}
\end{figure}


\newpage
\appendix

\section{Example Clinical Reasoning Outputs}
\label{sec:clinical_reasoning_examples}
Below, we show two example cases of clinical reasoning by both the panel and models. 

\subsubsection{Case VD-B-1-0001}

\noindent
\textbf{Primary diagnosis:} Acute hepatitis secondary to Hepatitis B infection \\
\textbf{Secondary diagnosis:} HIV

\paragraph{Panel clinical reasoning}
\begin{itemize}
    \item \textbf{Clinical picture:}
    \begin{itemize}
        \item Acute onset
        \item Hepatomegaly on examination makes an acute process more likely than a chronic one
    \end{itemize}
    \item \textbf{Positive Hepatitis B serology:}
    \begin{itemize}
        \item Positive IgM and surface antigen
        \item Negative anti-surface antibodies
    \end{itemize}
    \item Hepatocellular damage on liver function tests
    \item No evidence of chronicity on ultrasound
\end{itemize}

\paragraph{LLM model clinical reasoning}
The patient's history of HIV with a low CD4 count and high viral load, combined with a new onset of jaundice, elevated liver enzymes (ALT, AST, GGT), and predominantly conjugated hyperbilirubinemia, strongly points to acute liver injury. Crucially, the laboratory tests show positive HBsAg, positive HB core IgM, and a very high HBV viral load (2.78 million IU/mL), confirming an acute Hepatitis B infection. The abdominal ultrasound reported no obstructive cause for jaundice. The re-initiation of ART may also have triggered an immune reconstitution inflammatory syndrome (IRIS), leading to an HBV flare. The patient's acute jaundice and severe liver dysfunction are primarily caused by an acute Hepatitis B infection, exacerbated by advanced HIV disease and recent ART re-initiation.

\vspace{1em}

\noindent\textbf{Panel score of LLM clinical reasoning:} 5

\vspace{1em}
\hrule
\vspace{1em}

\subsubsection{Case VD-B-1-0138}

\noindent
\textbf{Primary diagnosis:} Ischemic stroke (cortical) - left middle cerebral artery
\\
\textbf{Secondary Diagnosis:} Hypertension, dyslipidemia

\paragraph{Panel clinical reasoning}
Acute onset weakness with right hemiplegia (arm $>$ leg), suggesting left middle cerebral artery involvement. Broca's aphasia suggests cortical involvement. CT scan shows no bleed; therefore, ischaemic stroke is most likely. Age, sex, raised blood pressure, and dyslipidaemia are risk factors for an ischaemic event.

\paragraph{Model clinical reasoning}
The patient's sudden collapse followed by acute onset of right-sided weakness and aphasia are hallmark signs of an acute stroke. The CT head images, particularly with contrast, would help rule out haemorrhage (I61.9), making ischaemic stroke (I63.9) the most likely diagnosis. Risk factors such as hyperlipidaemia and pre-diabetes support a cerebrovascular event. Metabolic acidosis (high anion gap, low bicarbonate) and elevated troponin are also noted. The patient presented with classic symptoms and neurological deficits indicative of an acute ischaemic stroke, supported by risk factors including hyperlipidaemia and pre-diabetic state.

\vspace{1em}

\noindent\textbf{Panel score of model clinical reasoning:} 5


\section{Scores on Data Subsets}
In this appendix, we provide tables showing the mean performance of the models on three key subsets of the data: (1) the Balanced dataset, (2) the Discordant dataset and (3) the subset of cases for which all models returned a diagnosis.

\begin{table*}[t]
\centering
\begin{subtable}[t]{\textwidth}
\centering
\renewcommand{\arraystretch}{1.2}
\begin{tabular}{p{0.13\linewidth} p{0.07\linewidth} p{0.07\linewidth} p{0.08\linewidth} p{0.07\linewidth} p{0.07\linewidth} p{0.07\linewidth} | p{0.07\linewidth} p{0.07\linewidth} | p{0.03\linewidth}}
\hline
\multicolumn{1}{c}{} &
\multicolumn{6}{l}{\textbf{RAW LLM JURY SCORES}} &
\multicolumn{2}{l}{\textbf{PANEL-CALIBRATED SCORES}} &
\\
\cmidrule(lr){2-7} \cmidrule(lr){8-9}
\textbf{Assessed Model} & \textbf{Dx} & \textbf{DDx} & \textbf{Clinical Reasoning} & \textbf{Patient Safety} & \textbf{$\mathbf{S_4}$ (RAW)} & \textbf{$\mathbf{S_3}$ (RAW)} & \textbf{$\mathbf{S_4}$ (ISO)} & \textbf{$\mathbf{S_3}$ (ISO)} & \textbf{Cases} \\
\hline

GPT-5.1 & \textbf{3.45} & \textbf{3.08} & \textbf{3.56} & \textbf{3.58} & \textbf{3.48} & \textbf{3.43} & \textbf{4.09} & \textbf{4.05} & 134 \\

Gemini 2.5 Pro & 3.44 & 2.85 & 3.47 & 3.56 & 3.43 & 3.37 & 4.07 & 4.04 & {\bf 139} \\

Gemini 3 Pro (Preview) & 3.36 & 2.70 & 3.54 & 3.51 & 3.39 & 3.29 & 4.05 & 3.99 & {\bf 139} \\

GPT-5 Mini & 3.25 & 2.92 & 3.40 & 3.46 & 3.32 & 3.27 & 4.03 & 4.00 & 133 \\

o3 & 3.39 & 2.82 & 3.43 & 3.47 & 3.37 & 3.30 & 4.03 & 4.00 & 134 \\

Claude 4.5 Sonnet & 3.29 & 2.74 & 3.43 & 3.43 & 3.32 & 3.24 & 4.03 & 3.98 & 94 \\

Grok 4.1 (Fast Reasoning) & 3.10 & 2.57 & 3.24 & 3.19 & 3.11 & 3.03 & 3.92 & 3.87 & 136 \\

Gemini 2.5 Flash & 3.06 & 2.54 & 3.14 & 3.24 & 3.09 & 3.03 & 3.90 & 3.86 &  {\bf 139} \\

o4-mini & 3.08 & 2.39 & 3.05 & 3.11 & 3.01 & 2.95 & 3.86 & 3.82 & 135 \\

Claude 4.1 Opus & 3.11 & 2.65 & 3.25 & 3.19 & 3.13 & 3.05 & 3.92 & 3.87 & 93 \\
\hline
Ward & 2.88 & 1.99 & -- & 2.70 & -- & 2.63 &  & 3.65 & 139 \\

\hline
\end{tabular}
\vspace{0.2em}
\caption{ Balanced Subset}
\label{tab:sub1}
\end{subtable}
\vspace{0.5em}
\begin{subtable}[t]{\textwidth}
\centering
\renewcommand{\arraystretch}{1.2}
\begin{tabular}{p{0.13\linewidth} p{0.07\linewidth} p{0.07\linewidth} p{0.08\linewidth} p{0.07\linewidth} p{0.07\linewidth} p{0.07\linewidth} | p{0.07\linewidth} p{0.07\linewidth} | p{0.03\linewidth}}
\hline
\multicolumn{1}{c}{} &
\multicolumn{6}{l}{\textbf{RAW LLM JURY SCORES}} &
\multicolumn{2}{l}{\textbf{PANEL-CALIBRATED SCORES}} &
\\
\cmidrule(lr){2-7} \cmidrule(lr){8-9}
\textbf{Assessed Model} & \textbf{Dx} & \textbf{DDx} & \textbf{Clinical Reasoning} & \textbf{Patient Safety} & \textbf{$\mathbf{S_4}$ (RAW)} & \textbf{$\mathbf{S_3}$ (RAW)} & \textbf{$\mathbf{S_4}$ (ISO)} & \textbf{$\mathbf{S_3}$ (ISO)} & \textbf{Cases} \\
\hline

GPT-5.1 & \textbf{3.28} & \textbf{3.01} & \textbf{3.44} & \textbf{3.42} & \textbf{3.34} & \textbf{3.28} & \textbf{4.01} & \textbf{3.97} & 154 \\

Gemini 3 Pro (Preview) & 3.23 & 2.59 & 3.38 & 3.36 & 3.25 & 3.16 & 3.98 & 3.93 & 159 \\
Gemini 2.5 Pro & 3.14 & 2.78 & 3.17 & 3.27 & 3.15 & 3.12 & 3.91 & 3.89 & {\bf  161} \\

GPT-5 Mini & 3.10 & 2.91 & 3.35 & 3.24 & 3.20 & 3.12 & 3.96 & 3.91 & 154 \\
Claude 4.5 Sonnet & 3.20 & 2.68 & 3.42 & 3.24 & 3.22 & 3.11 & 3.96 & 3.90 & 103 \\

o3 & 3.14 & 2.78 & 3.22 & 3.20 & 3.15 & 3.09 & 3.91 & 3.88 & 156 \\

Grok 4.1 (Fast Reasoning) & 2.88 & 2.56 & 3.07 & 2.95 & 2.93 & 2.84 & 3.80 & 3.75 & 152 \\

Gemini 2.5 Flash & 2.88 & 2.43 & 2.93 & 2.98 & 2.88 & 2.83 & 3.78 & 3.75 & {\bf 161} \\

o4-mini & 2.85 & 2.54 & 2.89 & 2.90 & 2.85 & 2.81 & 3.76 & 3.73 & 156 \\

Claude 4.1 Opus & 2.73 & 2.51 & 2.78 & 2.78 & 2.74 & 2.71 & 3.69 & 3.67 & 102 \\
\hline
Ward & 2.58 & 1.99 & -- & 2.39 &  & 2.39 & -- & 3.50 & 161 \\

\hline
\end{tabular}
\vspace{0.2em}
\caption{Discordant Subset}
\label{tab:sub2}
\end{subtable}
\vspace{0.5em}
\begin{subtable}[t]{\textwidth}
\centering
\renewcommand{\arraystretch}{1.2}
\begin{tabular}{p{0.13\linewidth} p{0.07\linewidth} p{0.07\linewidth} p{0.08\linewidth} p{0.07\linewidth} p{0.07\linewidth} p{0.07\linewidth} | p{0.07\linewidth} p{0.07\linewidth} | p{0.03\linewidth}}
\hline
\multicolumn{1}{c}{} &
\multicolumn{6}{l}{\textbf{RAW LLM JURY SCORES}} &
\multicolumn{2}{l}{\textbf{PANEL-CALIBRATED SCORES}} &
\\
\cmidrule(lr){2-7} \cmidrule(lr){8-9}
\textbf{Assessed Model} & \textbf{Dx} & \textbf{DDx} & \textbf{Clinical Reasoning} & \textbf{Patient Safety} & \textbf{$\mathbf{S_4}$ (RAW)} &  \textbf{$\mathbf{S_3}$ (RAW)} & \textbf{$\mathbf{S_4}$ (ISO)} & \textbf{$\mathbf{S_3}$ (ISO)} & \textbf{Cases} \\
\hline

GPT-5.1 & \textbf{3.30} & \textbf{3.04} & \textbf{3.46} & \textbf{3.48} & \textbf{3.38} & \textbf{3.32} & \textbf{4.03} & \textbf{4.00} & 195 \\

Gemini 3 Pro (Preview) & 3.27 & 2.67 & 3.44 & 3.41 & 3.30 & 3.20 & 4.01 & 3.95 & 195 \\

Gemini 2.5 Pro & 3.23 & 2.80 & 3.26 & 3.37 & 3.24 & 3.20 & 3.97 & 3.95 & 195 \\

Claude 4.5 Sonnet & 3.25 & 2.71 & 3.43 & 3.33 & 3.27 & 3.17 & 3.99 & 3.94 & 195 \\

o3 & 3.21 & 2.82 & 3.25 & 3.31 & 3.21 & 3.17 & 3.95 & 3.92 & 195 \\

GPT-5 Mini & 3.09 & 2.88 & 3.32 & 3.31 & 3.21 & 3.14 & 3.98 & 3.94 & 195 \\

Grok 4.1 (Fast Reasoning) & 3.02 & 2.63 & 3.22 & 3.11 & 3.07 & 2.98 & 3.89 & 3.84 & 195 \\

Gemini 2.5 Flash & 2.91 & 2.48 & 2.96 & 3.08 & 2.93 & 2.89 & 3.81 & 3.78 & 195 \\

Claude 4.1 Opus & 2.91 & 2.57 & 3.01 & 2.98 & 2.92 & 2.87 & 3.80 & 3.76 & 195 \\

o4-mini & 2.93 & 2.50 & 2.92 & 2.99 & 2.90 & 2.87 & 3.81 & 3.78 & 195 \\
\hline
Ward & 2.69 & 2.02 & -- & 2.54 & --  & 2.50 & -- & 3.58 & 195 \\

\hline
\end{tabular}
\vspace{0.2em}
\caption{195-case subset where all models returned a diagnosis }
\label{tab:sub3}
\end{subtable}
\vspace{0.5em}
\caption{Mean model scores (on scale 1-5) averaged across all three LLM Jury model scores and across all cases where a response was returned by the model (last column). Average $S_3$ score for LLMs vs ward are: 3.20 v 2.63 for (a) Balanced, 3.01 v 2.39 for (b) Discordant and 3.08 vs 2.50 for the subset of cases for which all models returned a diagnosis. The significant drop in all scores in the Discordant subset (b) and subset (c) compared with the Balanced subset (a) primarily appears to reflect greater diagnostic difficulty overall, as shown by lower scores for both LLMs and ward diagnoses. All cases excluded from subset (c) included more than 100 images, suggesting that case complexity is correlated with the number of MRI, CT scan and/or X-ray test results. Within the Discordant subset (b), expert adjudication still favoured LLM outputs over routine ward diagnoses on average, although the extra advantage relative to the Balanced subset (a) appears modest (difference of 0.61 vs 0.54). }
\label{tab:main}
\end{table*}

\end{document}